%% file: thesis.tex
\documentclass[11pt]{ucsddissertation}
\usepackage{amsmath}
\usepackage{tcolorbox}
\usepackage{subcaption}
\usepackage{mathrsfs}

\usepackage{array}
\usepackage{tabularx}
\usepackage{booktabs}
\usepackage{ragged2e}
\usepackage{microtype}
\usepackage[breaklinks=true,pdfborder={0 0 0}]{hyperref}
\usepackage{graphicx}
\usepackage{xcolor}
\usepackage{amsfonts}
\increasemargins{.1in}

\usepackage[english]{babel}

\usepackage{url}

\usepackage[super,sort&compress]{natbib}

\title{K-Inverse-RFM: A Modified RFM that Bridges the Gap to Neural Networks for Data-Corrupted Mathematical Tasks}
\author{Gil Moshe Pasternak}
\degree{Computer Science}{Master of Science}

\chair{Ramamohan Paturi}
\committee{Mikhail Belkin} 
\committee{Taylor Berg-Kirkpatrick}
\degreeyear{2025}

\begin{document}

\frontmatter
\maketitle{}
\makecopyright{}
\makesignature{}

\begin{dedication}
	\setsinglespacing{}
	\parindent0pt\parskip\baselineskip{}
	\vskip0pt plus.25fil
	\begin{center}
		
		To my incredible parents, who moved 12,500 miles throughout their lives\\
        to provide me the opportunity to submit this Thesis Today

		To my sister for being a friend to grow up with\\[1em]

		To my maternal grandparents for their infinite love, support, and friendship

        To my paternal grandparents for their love - I hope I've made you proud

        To my great grandfather Michael, for paving a path to follow 

        To Danielle, whose love pulls me through the harder days

        To Alec, Eric, John, Jesse, Alon, Assaf, Winkler: for endless support, advice, and laughter
		
	\end{center}
\end{dedication}


\tableofcontents
\listoffigures

\begin{acknowledgements}
    Getting a Master's Degree has been a profoundly different experience from the process of obtaining my Bachelor's. Whereas throughout my Bachelor's I learned to be accountable in the absolute for my results and performance, my Master's has been a story of collaboration and learning from others. In that light, there are several people without whom this process would not have reached fruition.

    First and foremost, I would like to express my sincere gratitude to my advisors, Mikhail Belkin and Ramamohan Paturi, without whom this thesis would not have been possible. Dr. Belkin—thank you for introducing me to the world of RFMs and AGOP, capturing my fascination with your lectures, and providing me with the mentorship and guidance that allowed me to contribute to this field. I am forever indebted to you for your support. Dr. Paturi—thank you for your guidance throughout my Master’s and for teaching me how to be a researcher. Your insights on the AI Tutor project and the Risk-of-Bias Benchmark project, your encouragement to keep up with the latest literature, and your direction in shaping this very thesis have been invaluable. I cannot imagine pursuing a research career without your mentorship.

    Much of the work presented in this thesis has also been guided by Neil Mallinar, whose endless patience and willingness to help I deeply appreciate. Notably, Neil ideated the first version of K-Inverse I tried, and I am grateful to have gotten the opportunity to build on top of his work. I am also thankful for the contributions of Leon Bergen, Andre Wang, and the rest of the LEI group. Dr. Bergen — thank you for helping me navigate my early research efforts, introducing me to the broader research landscape, and guiding me through my first start-to-finish research project. Andre Wang and the rest of LEI — thank you for placing your trust in me and for your collaboration on the Risk-of-Bias Benchmark project. Further, I would like to express my gratitude to Hannah Carter, James Talwar, Seth Levine, and Colin Jemmott — my first research mentors. Your guidance is the reason I have continued to pursue research. Dr. Carter — thank you for your mentorship and for helping me plan my graduate studies as a whole. James — I greatly appreciate your friendship and your endless wellspring of advice, which has helped me accomplish far more in two years than I ever could have imagined. Seth — your mentorship (and friendship), guidance, and encouragement gave me the creative freedom to grow into my own as a researcher. You played a major role in shaping my passion for this profession, and for that, I am incredibly grateful. Professor Jemmott — thank you for believing in me when I was an 18-year-old with no experience whatsoever. Without your course and the Likelihood research project, I don’t know that I would be pursuing research, a Master’s degree, or even a career in artificial intelligence at all.
    
    None of this work would be possible without my family. Mom and Dad - you escaped a crumbling Soviet Union and uprooted your lives twice to move countries to find something better for you and your children. You moved thousands miles away from parents, relatives, and friends for us, and toiled endlessly to make my American dream possible. You invested obsessively in educating me  both morally and intellectually. Needless to say, I would not be here without you. To Yael - thank you for being a great friend through it all, I really appreciate you. To my maternal grandparents - thank you for being my daily first call. Our very regular conversations and endless laughter has made me a fundamentally happier person and motivated me to work towards horizons I never thought possible. To my paternal grandparents - thank you for all your love and support. May your memory be a blessing. 

    Finally, no acknowledgment would be complete without recognizing the love and friendship that have enriched my life. Danielle — thank you for your ceaseless love, positivity, and encouragement throughout this thesis journey. Your strength, radiance, and love of life inspire me and you have been my  motivation on those days when nothing seems to go quite right. I appreciate you so much. Alec — I massively appreciate your friendship over the past decade. Looking back, I cannot name anything I ever accomplished where you did not somehow play a part, whether it was teaching me how to write, helping me get STEMLearn off the ground while you were working full time, or helping me make tough life decisions. Eric — beyond your work ethic pushing me to actually complete this thesis during our daily joint work sessions or your partnership helping turn STEMLearn into reality, you have been one of the most reliable, consistent, and trustworthy people in my life for over eight years. I can’t thank you enough. John — thank you for eight years of incredible friendship, countless shared laughs over things no one else would understand, and for your infinite kindness and dependability. Alon - thank you for being the world's best housemate for the entire duration of my Master's degree, being a willing ear for any research discussion, and generally being an awesome friend. Jesse - thank you for always being a willing ear for both technical and non-technical (life) problems, helping me debug code on countless occasions, and generally for 8 years of really great friendship.

There are countless others who have profoundly impacted my life over the past two years. Though I cannot name everyone here, I want to extend my gratitude to Winkler, Assaf, Daniel, Alex, Eugene, Mendy, Dr. Fung, the Maxes, Varun, Upo, and many more—thank you for all you've done.

This thesis is a product of the mentorship, support, and friendship of so many people, and I am beyond grateful to each and every one of you.

\end{acknowledgements}

\begin{dissertationabstract}
Recursive Feature Machines (RFMs) are a class of kernel machines that utilize the Average Gradient Outer Product (AGOP) as a mechanism for feature learning. They have been shown to effectively replicate the learning dynamics and feature representations of Feedforward Neural Networks (FNNs) across various settings. However, despite comparable capacity for feature learning and the similarities in the features they acquire, RFMs exhibit significantly lower performance than neural networks in certain data-dependent scenarios. In this work, we investigate these limitations in mathematical problems. As a solution, we introduce a remarkably effective transformation applied to the training labels which promotes learning in noisy, complexly represented, and class-imbalanced data. This simple yet powerful adjustment enables RFMs to close the performance gap with FNNs and, in some cases, even surpass them.	
\end{dissertationabstract}

\mainmatter{}

\include{1_introduction}

\include{2_kernels_agop_rfm}
\include{3_the_gap}

\include{4_k_inverse}
\include{5_conclusions_and_future_avenues}

\appendix

\include{A1_other_interesting_avenues_observed}

\backmatter{}
\bibliographystyle{unsrtnat} 
\bibliography{refs}

\end{document}

%% file: 1_introduction.tex
\chapter{Introduction}
Neural networks are among the most powerful and widely used learning algorithms today, with applications spanning language processing, computer vision, and biology. A fundamental aspect of their success is their ability to learn non-linear features - often in a lower-dimensional subspace. This feature-learning mechanism provides a significant advantage over classical methods such as linear regression, decision trees, and boosting, which typically operate with fewer parameters and lack an implicit bias for discovering complex feature interactions.

Recent research suggests that integrating a feature-learning mechanism called the Average Gradient Outer Product (AGOP) with kernel regression can help bridge the gap between classical methods and neural networks. The resulting models, known as Recursive Feature Machines (RFMs) \cite{originalrfm2023}, have been shown to match or exceed neural network performance on tabular data benchmarks. They also replicate neural network feature representations with high correlation and exhibit key neural learning phenomena, including the Lottery Ticket Hypothesis and Grokking. An empirically validated Neural Feature Ansatz suggests that RFMs tend to approximate the Neural Feature Matrices of a neural network.

In this work, we conduct a thorough investigation of the differences between neural networks and RFMs for mathematical problems such as modular arithmetic and greatest common divisor. Through a broad set of experiments, we observe a gap between RFMs and neural networks in mathematical tasks where the data is corrupted with label noise, has an uneven label distribution, or is represented in complex ways. We provide a comprehensive review of these results in Chapter 3. 

Then, in Chapter 4, we formulate and justify a solution which we call the "K-Inverse-RFM", which is a  modification of the RFM that addresses these problems. The "K-Inverse-RFM" consistently bridges some or all of the gap between RFMs and neural networks in these contexts, at times outperforms neural networks, and can be easily trained on a CPU. We also provide comprehensive results on all tasks and insight as to when the K-Inverse-RFM might be successful.

Finally, between Chapter 5 and the Appendix, we provide an overview of potential future avenues and other interesting RFM/neural net related observations that we have gathered from our experiments. In totality, our contributions are as follows:

\begin{itemize}
    \item We conduct a thorough investigation of the performance of RFMs on various discrete mathematical tasks.
    \item We triangulate and understand various gaps between neural network performance and the performance of a RFM.
    \item We propose a modification of the RFM to address these shortcomings and provide a thorough comparison of its performance.
    \item We explain other interesting neural network phenomena discovered in our experiments, showing that the entire performance of a neural network could at times be surpassed using just first layer features. 
\end{itemize}

%% file: 2_kernels_agop_rfm.tex
\newtheorem{definition}{Definition} 
\newtheorem{algorithm}{Algorithm}

\chapter{Relevant Background}
\section{Kernels and Kernel Ridge Regression}
We begin by providing an introduction to kernels and kernel ridge regression, the latter of which is used as a predictive model by our RFM. Kernel methods are a fascinating area of study and can be covered in great depth \cite{liang2016kernel}, but for the sake of brevity this work covers only those intuitions, definitions and properties which are directly relevant to the RFM and our experiments.

We begin by motivating the need for a kernel. Assume we have a learning task where our labels are a nonlinear function of our inputs. We define $X = \{x_i\}_{i=1}^{n} \in \mathbb{R}^{n \times d}$ to be our dataset and Y $\in \mathbb{R}^{n}$ to be our label vector.

Linear methods, commonly of the form $f(x) = \langle w, x\rangle $ where w is a learned weight vector, would always fail in performing this task. However, we can perform a simple trick to attain a form of regression that might work. We can reframe our function as $f(x) = \langle w, \phi(x)\rangle $, where $\phi:\mathbb{R}^d \xrightarrow[]{} \mathbb{R}^h$ is some high dimensional nonlinear projection of $x \in \mathbb{R}^d$, such as the map $[x_1, ... x_n] \xrightarrow{} [x_1^2, \dots, x_n^2, \sqrt2x_1x_2, ...\sqrt2x_1x_b, \sqrt2x_2x_3, \dots, \sqrt2x_2x_b, \dots, \sqrt2x_{b-1}x_b]$. Now, as a result of our higher dimensional projection $\phi$ that includes second order features, we are able to model second degree polynomials.

There is only one problem with this approach - going from a lower dimension $d$ to a higher dimension $h$ now means that our dot product $\langle w, \phi(x)\rangle $ is $O(h)$ to compute, which could be expensive. Luckily, a short computation allows us to reframe this in a computationally efficient manner. Starting with the MSE of our predictor:
$$MSE = \frac{1}{n}\sum_{i=1}^{n}\frac{1}{2}(y^{(i)} - w^\top\phi(x^{(i)}))^2$$
Means that our gradient with respect to w is:
\[
\frac{\partial \mathcal{L}}{\partial w} = \frac{1}{n}\sum_{i=1}^{n} (y_i - w^T \phi(x_i)) (-\phi(x_i))
\]
Which, if we start w at 0 and update using gradient descent, this indicates that our final weight vector w is a sum of scalars multiplied by $\phi(x^{(i)})$, so then we reformulate w as:
$$w = \sum_{i=1}^{n}\alpha_i\phi(x^{(i)})$$
This then implies:
$$f(x) = \langle \sum_{i=1}^{n}\alpha_i\phi(x^{(i)}), \phi(x) \rangle \xrightarrow{}$$
$$f(x) = \sum_{i=1}^{n}\alpha_i \langle \phi(x^{(i)}), \phi(x)\rangle$$

If we can find a way to compute $\langle \phi(x^{(i)}, \phi(x)\rangle$ efficiently for our selected function, we have successfully computed f(x) efficiently. Fortunately, it can be verified that $\langle \phi(x^{(i)}), \phi(x)\rangle = \langle x^{(i)}, x \rangle ^2$, which can be computed in $O(d)$ time since $x \in \mathbb{R}^d$. Our final computation for the entire sum $\sum_{i=1}^{n}\alpha_i \langle \phi(x^{(i)}), \phi(x)\rangle$ is $O(dn)$, whereas the naive computation would have been $O(d^2n)$.

This finding of an efficient alternative computation for a high-dimensional nonlinear projection is called the kernel trick, and the computation of the high-dimensional dot product itself is called a \textbf{kernel} and notated $k(x_i, x_j)$. We also often leverage kernel matrices, which are matrices of kernel computations defined as $K(X, U)_{i,j} = k(x_i, u_j)$.

In this work, we apply kernels in the context of \textbf{kernel ridge regression}, a non-parametric model which - given a Hilbert space $\mathcal{H}$ - we use to solve the optimization problem:

$$min_{f \in \mathcal{H}} \sum_{i=1}^n \frac{1}{2}\|f(x_i) -y_i\|^2_2 + \frac{\lambda}{2}\|f\|^2_\mathcal{H}$$

While the derivation itself is left out for the sake of concision \cite{liang2016kernel}, the very elegant representer theorem \cite{wahba1990spline} guarantees that the solution to this problem is both in the span of the data points and available in closed form:

$$f_{best}(x) = \sum_{i=1}^n \alpha_i k(x, x_i)$$
$$\alpha_i = (K(X,X) + \lambda I)^{-1}Y$$

Finally, we'll also occasionally refer to Mahalanobis kernels thorughout this work, which are just kernels leveraging the Mahalanobis distance metric, defined as:
\[
D_M(\mathbf{x}, \mathbf{x'}) = \sqrt{ (\mathbf{x} - \mathbf{x'})^T \sum^{-1} (\mathbf{x} - \mathbf{x'})}
\]
In the formulation above $\sum^{-1}$ refers to the inverse of the sample covariance matrix. In our formulations, we replace $\sum^{-1}$ with the AGOP matrix M, which we proceed to explain in the next section.


\section{AGOP and the Neural Feature Ansatz}
Having completed our discussion of kernels and kernel regression, we now elucidate the feature learning mechanism of our model, which we call the Average Gradient Outer Product (AGOP) Matrix \cite{originalrfm2023}. We define this matrix as follows: 

\begin{definition}
Let f be a predictor where \( f: \mathbb{R}^d \to \mathbb{R}^c \) has c outputs $[f_0,..., f_{c-1}(x)]$ \newline
And set the Jacobian J at a given point x' as $J_{u,v}(x) = \frac{\partial{f_u(x')}}{\partial{x_v}}$.
Then if f is trained on a dataset $\{x_j\}_{j=1}^{n}$ we define the \textbf{Average Gradient Outer Product} as 
$$G(f; \{x_j\}_{j=1}^{n}) = \frac{1}{n}\sum_{j=1}^{n}J(x_j) J(x_j)^T \in \mathbb{R}^{d \times d}$$
\end{definition}

To understand this matrix, we consider a single entry $G(f; \{x_j\}_{j=1}^{n})_{u,v}$, which we'll notate $G_{u,v}$ in shorthand. $G_{u,v}$ is then equivalent to $\frac{1}{n}\sum_{i=1}^{n}J(x_i)_u J(x_i)_v^T$, which is the average similarity between the gradients with respect to input feature $u$ and gradients with respect to input feature $v$. In a sense, this is similar to a covariance matrix with gradients - it quantifies similarity in how various input features change the output. A diagonal entry, on the other hand, would compute as: $G_{u,u} = \frac{1}{n}\sum_{i=1}^{n}J(x_i)_u J(x_i)_u^T = \frac{1}{n}\sum_{i=1}^{n}\||J(x_i)_u||_2^2$, which is the average squared norm of the gradient of an input feature across the dataset, indicating the magnitude of the feature's gradient. This can also be thought of as the feature's ``importance".

We claim that multiplication of the input features by the AGOP is a feature learning mechanism which amplifies important features and shrinks dependence on ``less important ones". To see this, assume that the first feature is irrelevant in impacting the output of a function f (all gradients of the output with respect to the first feature are 0 across input examples $x_1, x_2, ... x_n$), so that the Jacobian is structured as follows: 
$$
\begin{bmatrix}
0 & 0 & 0 & \cdots & 0 \\
J_{2,1} & J_{2,2} & J_{2,3} & \cdots & J_{2,c} \\
J_{3,1} & J_{3,2} & J_{3,3} & \cdots & J_{3,c} \\
\vdots & \vdots & \vdots & \ddots & \vdots \\
J_{d,1} & J_{d,2} & J_{d,3} & \cdots & J_{d,c}
\end{bmatrix}
$$

Then the AGOP would similarly maintain the structure:
$$
\begin{bmatrix}
0 & 0 & 0 & \cdots & 0 \\
0 & G_{2, 2} & G_{2,3} & \cdots & G_{2,d} \\
0 & G_{3,2} & G_{3,3} & \cdots & G_{3,d} \\
\vdots & \vdots & \vdots & \ddots & \vdots \\
0 & G_{d,2} & G_{d,3} & \cdots & G_{d,d}
\end{bmatrix}
$$

Hence, multiplying our input examples $x_1, x_2, ... x_n \in \mathbb{R}^d$ by this AGOP matrix would zero out the first feature such that it is not considered. Likewise, if the gradient of a function with respect to an input feature is larger, its corresponding row in the AGOP matrix will have larger values, and the feature will be amplified. Thus, one can view the AGOP as a feature learning mechanism that amplifies features which affect the output of a function and de-emphasizes features which don't. For a learned predictor $f(x)$, this means the amplification of the features which were learned to affect the predictors output. 

Leveraging this perspective, we can see the eigenvalues of the AGOP matrix as the most essential features of the input space, and the AGOP as a mechanism which amplifies relevant input features \cite{originalrfm2023, grokkingpaper} and improves our ability to capture functions of a low-dimensional subspace of our inputs.

Previous work has related the AGOP to neural networks with the empirically validated and theoretically motivated \textbf{Neural Feature Ansatz} \cite{originalrfm2023}, which states the following:

\begin{tcolorbox}[colframe=black, colback=white, sharp corners=south]
    \textbf{Neural Feature Ansatz:} For any layer $l$ of a fully trained neural network, given fully trained weights $W_l$, the form $W_l^TW_l$ is highly correlated to the AGOP matrix with respect to the layer input of $l$.
\end{tcolorbox}

Essentially, this empirically validated Ansatz states that we can often approximate a function of the weights of a fully trained network without multiple passes of backpropagation, thus showing the great advantage of using the AGOP mechanism.

\section{Recursive Feature Machines}
We note that the AGOP mechanism introduced in the previous section is quite general and can be applied as a feature learning mechanism to any differentiable predictor. One appealing option is kernel regression algorithm presented earlier, as it has no feature learning mechanism of its own and yet is able to learn complex, nonlinear, high dimensional interactions. In this light, Radhakrishnan et. al \cite{originalrfm2023} introduces the \textbf{Recursive Feature Machine (RFM)}, defined as follows \cite{grokkingpaper}: 
\begin{algorithm}
Let $(X,y) \in \mathbb{R}^{n \times d} \times \mathbb{R}^d$ be a set of training samples. We denote the n samples of X as $\{x_j\}_{j=1}^{n}$, $M_0 \in \mathbb{R}^{d \times d}$ to be an initial symmetric positive-definite matrix, and $K(\cdot, \cdot; X)$ is a Mahalanobis kernel. Then for $t \in [T]$, we iterate the following formulation to train our Recursive Feature Machine:
\begin{align}
    \alpha_t &=  K(X, X; M_t)^{-1}y && \text{\textbf{(Predictor Training)}} \\
    f_t(x) &= K(x, X; M_t) \alpha_t \\
    M_{t+1} &= G(f_t)^s && \text{\textbf{(Feature Learning)}} 
    \label{RFM}
\end{align}

Where s is a fractional matrix power, and $K(X,X;M)$ $\in \mathbb{R}^{n\times n}$ is the kernel matrix defined with entries $K(X,X;M)_{i,j}$ = $K(x_i, x_j, M) $ for i, j $\in$ $[n]$. As our kernel K, we use the three formulations listed below:

\begin{align*}
    K_{\text{Laplace}}(x, x';M) &= \exp\left( -\gamma\|x - x'\|_M \right) && \text{(Laplace kernel)} \\
    K_{\text{Gaussian}}(x, x';M) &= \exp\left( -\frac{\|x - x'\|_M^2}{L} \right) && \text{(Gaussian kernel)} \\
    K_{\text{Quadratic}}(x, x';M) &= (x^\top Mx')^2 && \text{(Quadratic kernel)}
\end{align*}

Where L is a bandwidth parameter, $\gamma > 0$, and $\|z\|_M = z^\top Mz$.
\end{algorithm}

It is worthwhile to note that each of these kernel formulations transforms its features using the AGOP matrix M.

RFMs have been previously been shown to perform at state of the art levels on tabular data \cite{originalrfm2023}, steer LLM concepts \cite{llmsteering}, recover low rank matrices \cite{rfmlowrankmatrices}, and reproduce various neural network phenomena such as Grokking \cite{grokkingpaper} and Deep Neural Collapse \cite{agopdeepneuralcollapse}.
\section{Grokking Modular Arithmetic}
Previous work \cite{grokkingpaper} has shown that RFMs can be used to replicate both neural network performance and the phenomena of ``Grokking" in modular addition, subtraction, multiplication and division tasks. This work also discussed how the phenomena of Grokking in RFMs was attributable to the gradular learning of block-circulant features, which upon materializing were used to implement the Fourier Multiplication Algorithm and reach perfect test set performance. 

A significant portion of this work will expand upon these results, showing that a performance gap between RFMs and neural networks begin to emerge if we modify/corrupt our training data in various ways, and simultaneously that this gap is bridgeable with a modification to RFM formulation. We begin by establishing this gap in the following section.

%% file: 3_the_gap.tex
\chapter{The RFM-NN Gap}
Mallinar et. al \cite{grokkingpaper} tested the performance of RFMs across a suite of modular arithmetic tasks. Representing the data in one-hot encodings and splitting off a fraction $f$ of all possible unique examples to serve as the task training set, the paper found that RFMs properly reproduced the perfect performance of neural networks, their features, and their grokking phenomena. We share our reproduction of a few result plots below:
\begin{figure}[h]
    \centering
    \begin{subfigure}{0.4\textwidth}
        \centering
        \includegraphics[width=\linewidth]{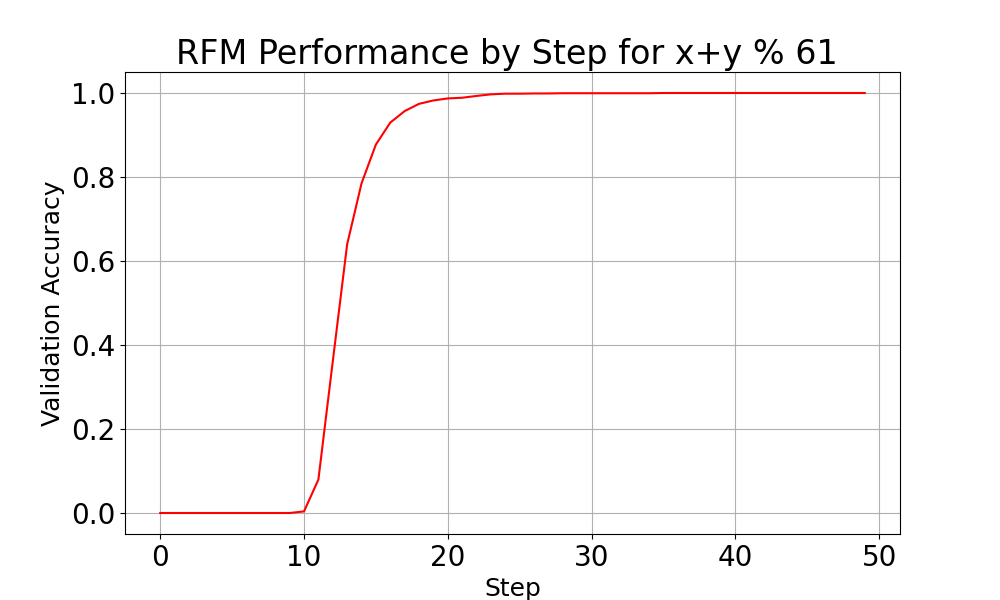}
    \end{subfigure}
    \begin{subfigure}{0.4\textwidth}
        \centering
        \includegraphics[width=\linewidth]{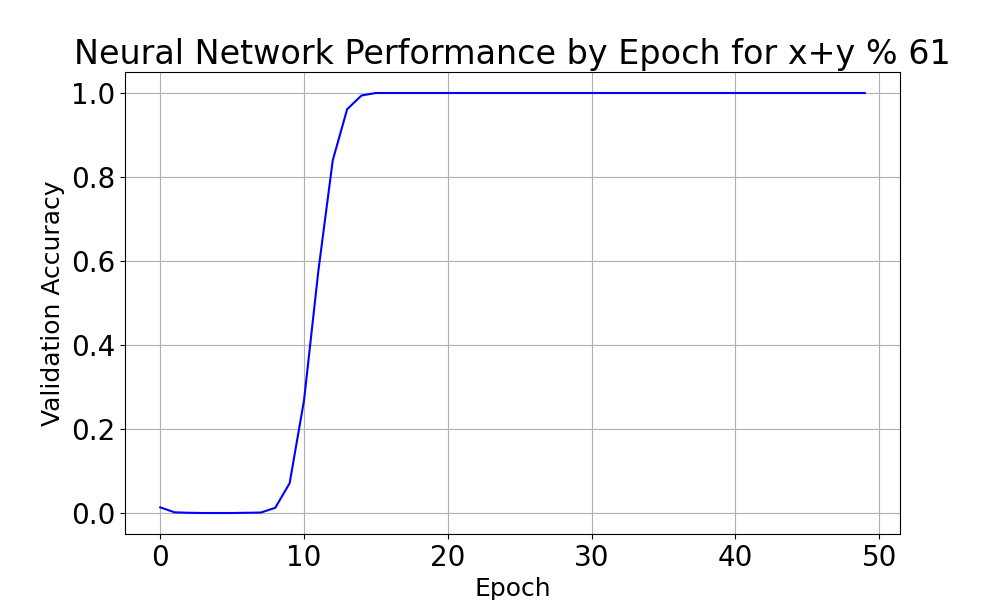}
    \end{subfigure}
    
    \label{fig:rfm_vs_nn_mod_addition}
     \begin{subfigure}{0.4\textwidth}
        \centering
        \includegraphics[width=\linewidth]{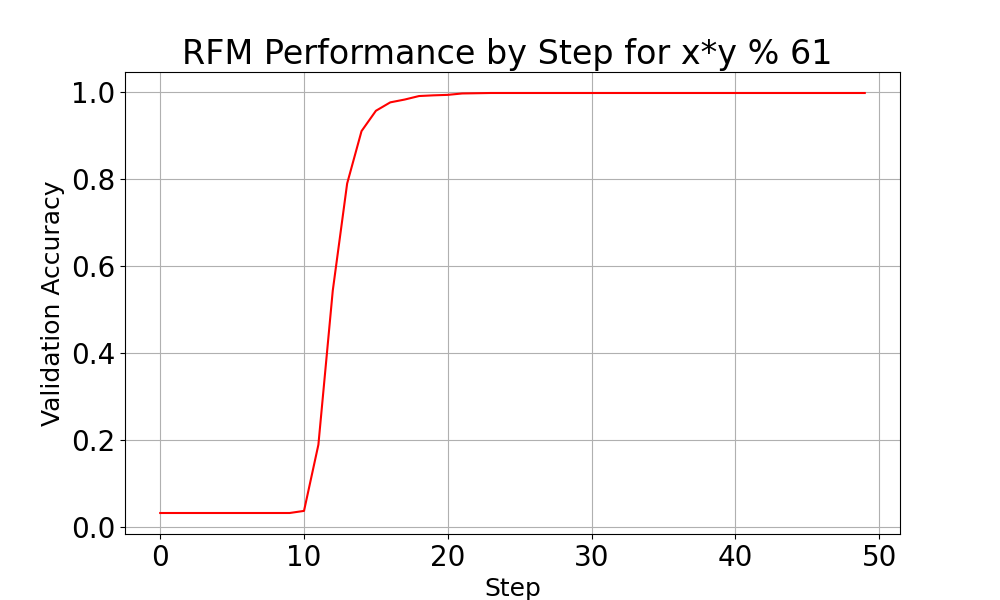}
    \end{subfigure}
    \begin{subfigure}{0.4\textwidth}
        \centering
        \includegraphics[width=\linewidth]{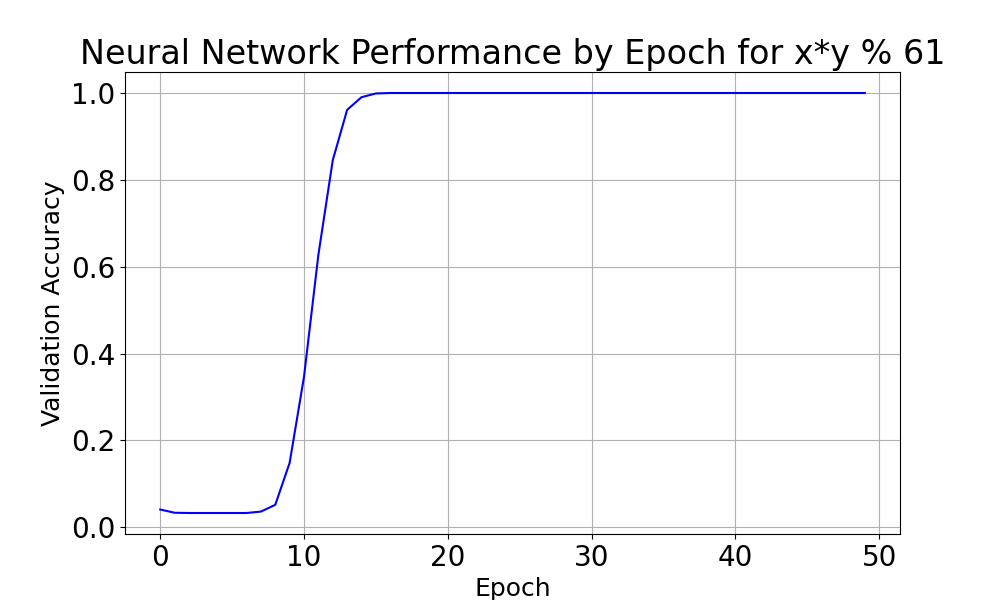}
    \end{subfigure}
    \caption{RFM vs. neural network test set performances by training step for modular addition (top, p=61) and multiplication (bottom, p=61). }
\end{figure}

In our experiments, we attempt a plethora of variations upon these and other tasks in comparison of neural networks and RFMs. In this light, we uncover three interesting phenomena:

\begin{enumerate}
    \item Neural networks perform far better than RFMs as label noise is introduced, scaling at a much better rate comparably. RFM performance drops off drastically.
    \item Neural networks do a better job than RFMs at dealing with class-imbalanced data: the performance of RFMs drops off whereas neural network performance remains steady.
    \item Neural networks do better than RFMs with complex representations, such as a Chinese Remainder Theorem based representation.
\end{enumerate}

\section{Label Noise}
Modular arithmetic tasks have been shown to be an arena of equivalent performance between neural networks and RFMs. To see if this phenomena extends only to clean data, we test whether the RFMs and neural networks will respond similarly to label noise. 

We set up the experiment as follows: we pre-select a prime modulo $p=61$. We test performance on five tasks: modular addition, modular subtraction, modular multiplication, modular division, and GCD. For each task, we assign half of the data to a training set, and the other half to a testing set. All data is one hot encoded. Leveraging this setup, we test the robustness of RFMs and neural networks to training set label noise, ranging from $2\%-32\%$ at increments of $2\%$. We present results below:
\begin{figure}[h]
    \centering
    \begin{subfigure}{0.4\textwidth}
        \centering
        \includegraphics[width=\linewidth]{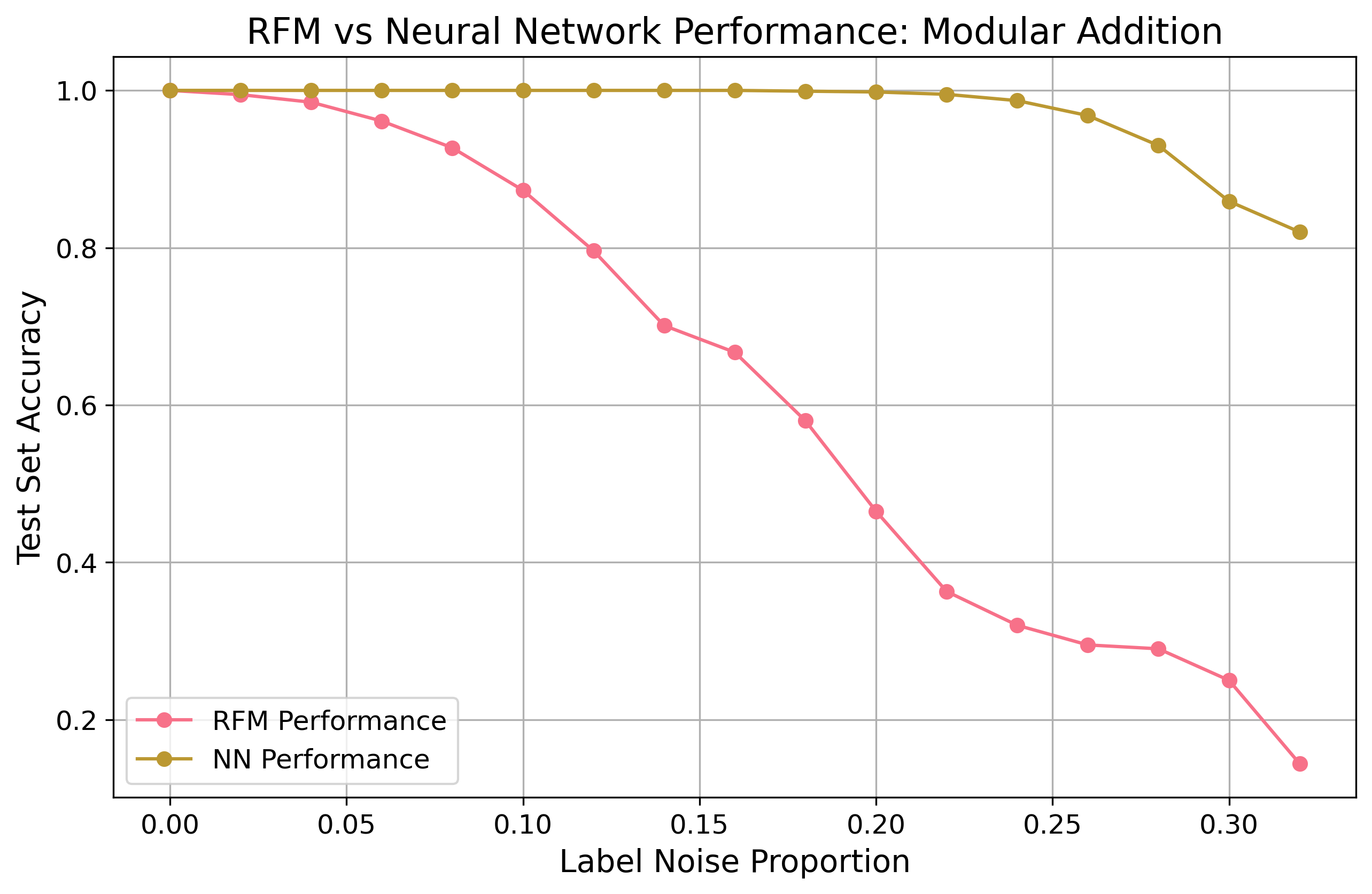}
    \end{subfigure}
    \begin{subfigure}{0.4\textwidth}
        \centering
        \includegraphics[width=\linewidth]{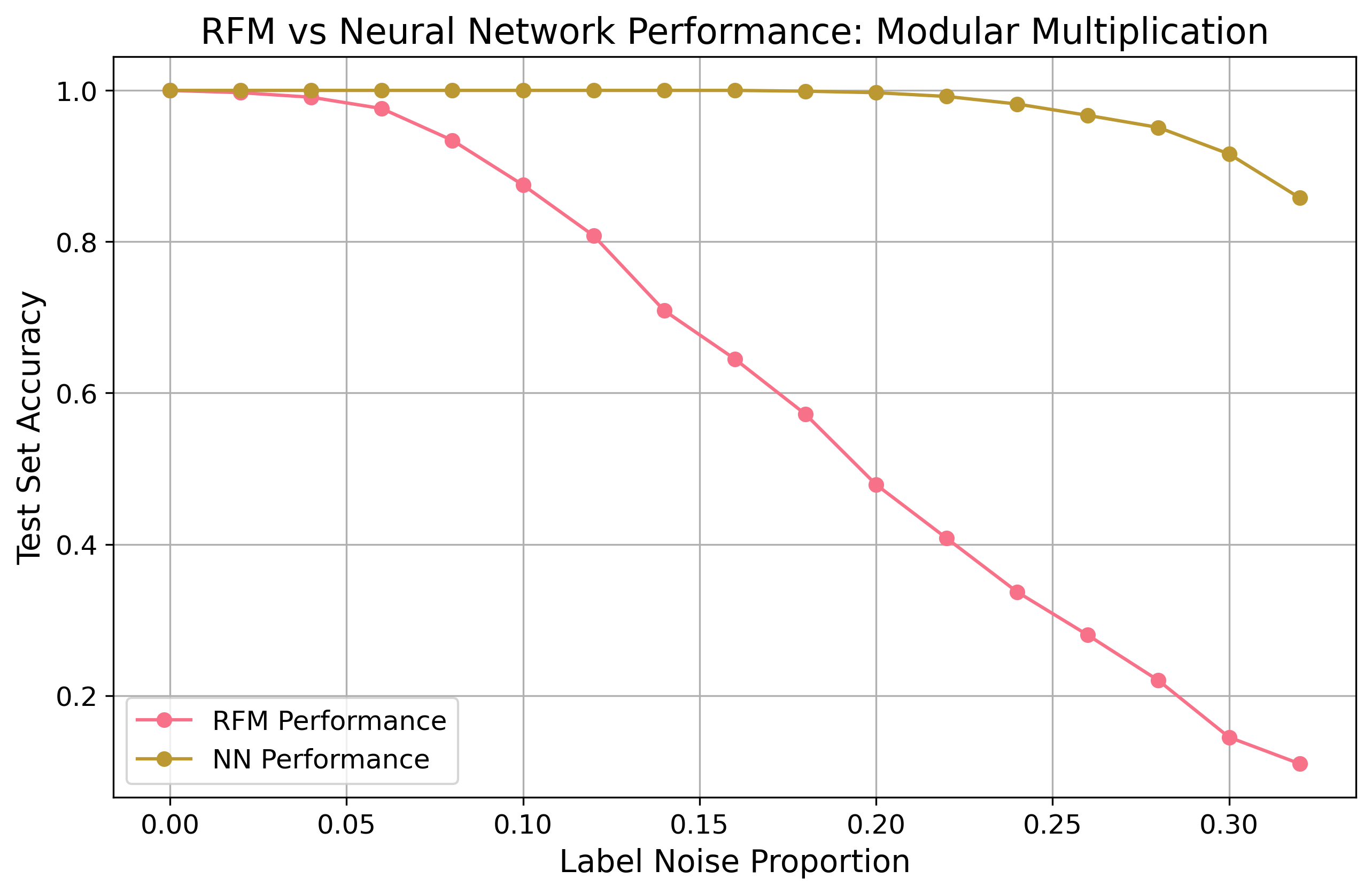}
    \end{subfigure}
    \caption{RFM vs. neural network test set Performances by training step for modular addition (left, p=61) and multiplication (right, p=61). }
    \label{rfm_nn_label_noise_comparison}
\end{figure}

What we observe in Figure \ref{rfm_nn_label_noise_comparison} is a notable disparity: neural networks scale far more robustly to label noise as opposed to RFMs. While neural networks manage to keep their testing accuracy above 80\%, RFMs quickly decay to performance that is essentially an order of magnitude worse. At $32\%$ label noise, the disparity between RFM and neural network performance on the test set is a vast $68\%$.

A natural question arises as to whether this arises from a superior ability of neural networks to learn features. We test this hypothesis in two ways. 

Firstly, we take the post-activation first layer features that the neural network learned at two label noise levels ($16\%$ and $32\%$) and apply a Laplace kernel to them to create a predictor. This allows us to see if the gap is defined by the superior feature learning capabilities of the first layer of the neural network in noisy settings. This will also provide evidence as to whether increased label noise degrades the feature quality of the network. 

Secondly, we take random circulant features \cite{grokkingpaper}, which are random features pre-constrained to a circulant structure utilized by modular arithmetic tasks. This allows us to see how much enforcing an implicit feature structure will help the RFM learn in noisy setting. A significantly improved performance in this context might indicate that the RFM fails to learn the necessary feature structure in noisy settings.

\begin{figure}[h]
    \centering
    \begin{subfigure}{0.4\textwidth}
        \centering
        \includegraphics[width=\linewidth]{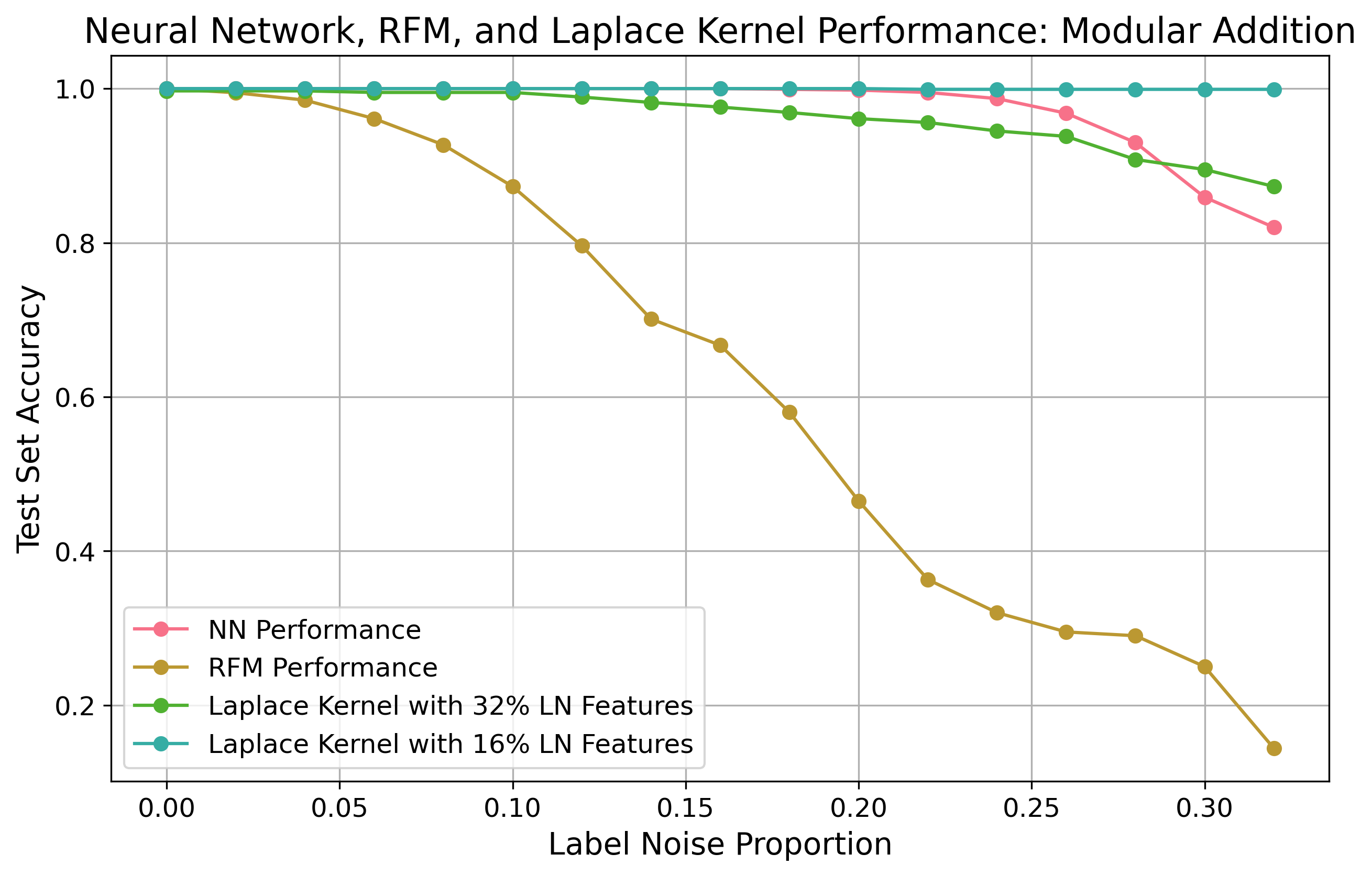}
    \end{subfigure}
    \begin{subfigure}{0.4\textwidth}
        \centering
        \includegraphics[width=\linewidth]{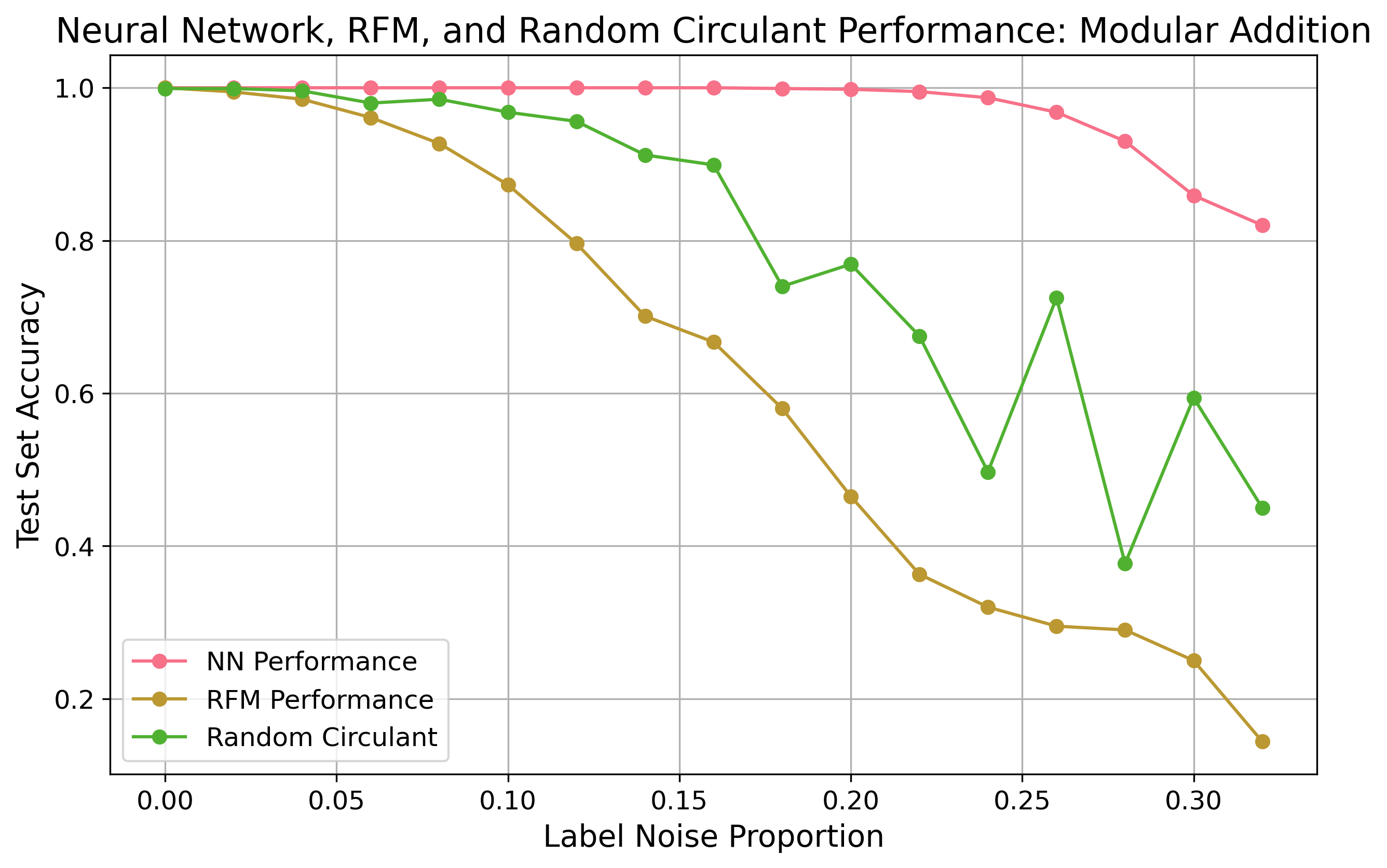}
    \end{subfigure}
    \caption[Comparison of various methods for modular addition with noisy training data...]{Comparison of various methods for modular addition with noisy training data. \textbf{Left:} Plot Comparing the test set performance of RFMs with neural networks and Laplace kernel regression trained on neural network features at $16\%$ / $32\%$ Label Noise. The kernel regression trained on the neural features outperforms other methods, indicating RFMs primary issue in noisy settings is feature learning. \textbf{Right:} kernel regression equipped with structured random circulant features significantly outperforms normal RFMs, suggesting RFMs fail to learn essential feature structure in noisy settings. }
    \label{rfm_nn_feature_learning_gap}
\end{figure}

The results, shown in Figure \ref{rfm_nn_feature_learning_gap}, are very strong evidence that the gap between Laplace kernel regression and neural networks lies primarily in the post-activation features learned by the first layer. When applying features learned by the network at $16\%$ label noise, the model achieves close to universally perfect accuracy at all levels of label noise, outperforming the neural network. This serves to demonstrate that the training label noise isn't tricking the RFM's predictive mechanism (kernel regression) but rather its feature learning mechanism (AGOP). Interestingly enough, the neural network's features also decay in quality with growing label noise: while the features learned at $32\%$ label noise still outperform those of the RFMs by a wide margin, they fare worse than the features learned at $16\%$ label noise across the board. Another interesting observation is that the Laplace kernel prediction head, when trained on neural network features, actually significantly outperforms the the remaining neural network layers.

Our results for random circulant features show much of the same. Forcing our predictive head to use random features of the correct form allows it to learn the task far more effectively than before. Holistically, this strongly suggests that increasing label noise corrupts the abilities of models to learn features, though much less so in a neural network than in an RFM.

\section{Imbalanced Data}
\subsection{Input and Output Imbalances}
The original results in Mallinar et. al \cite{grokkingpaper} show equivalent performance and improved speed of training for modular arithmetic tasks given a dataset created by $\{1,2,...p\} \times \{1,2,...p\}$, meaning each pair in the dataset appears exactly once. In this section, we check the results of resampling both the input and output distribution with various imbalances to see if this might cause a NN-RFM gap to emerge.

We set up two general experiments, evaluated on modular arithmetic and GCD as reported earlier. In the first experiment, we select a random inpur pair weighting condition: we adjust the weight of pairs where both elements are divisible by 3. We then take the dataset $\{1,2,...p\} \times \{1,2,...p\}$ and resample it with replacement, in each run changing the weight given to these divisible-by-3 pairs. We compare our RFM, trained for 200 iterations with a quadratic kernel to a neural network with two hidden layers and a quadratic activation function trained for 500 epochs. Both training times were normally sufficient for convergence (models reached a training accuracy of 1).

\begin{figure}[h]
    \centering
    \begin{subfigure}{0.4\textwidth}
        \centering
        \includegraphics[width=\linewidth]{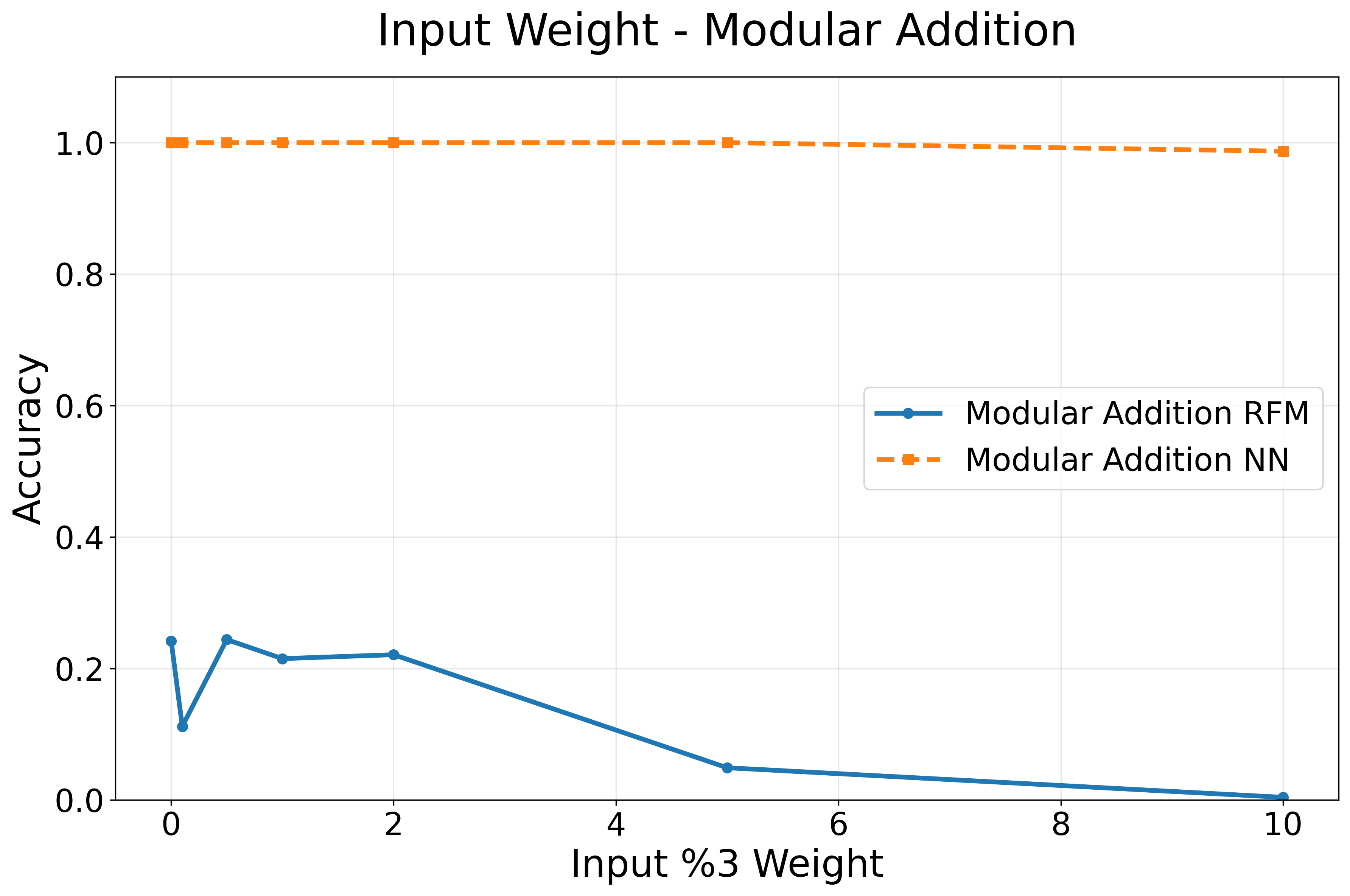}
    \end{subfigure}
    \begin{subfigure}{0.4\textwidth}
        \centering
        \includegraphics[width=\linewidth]{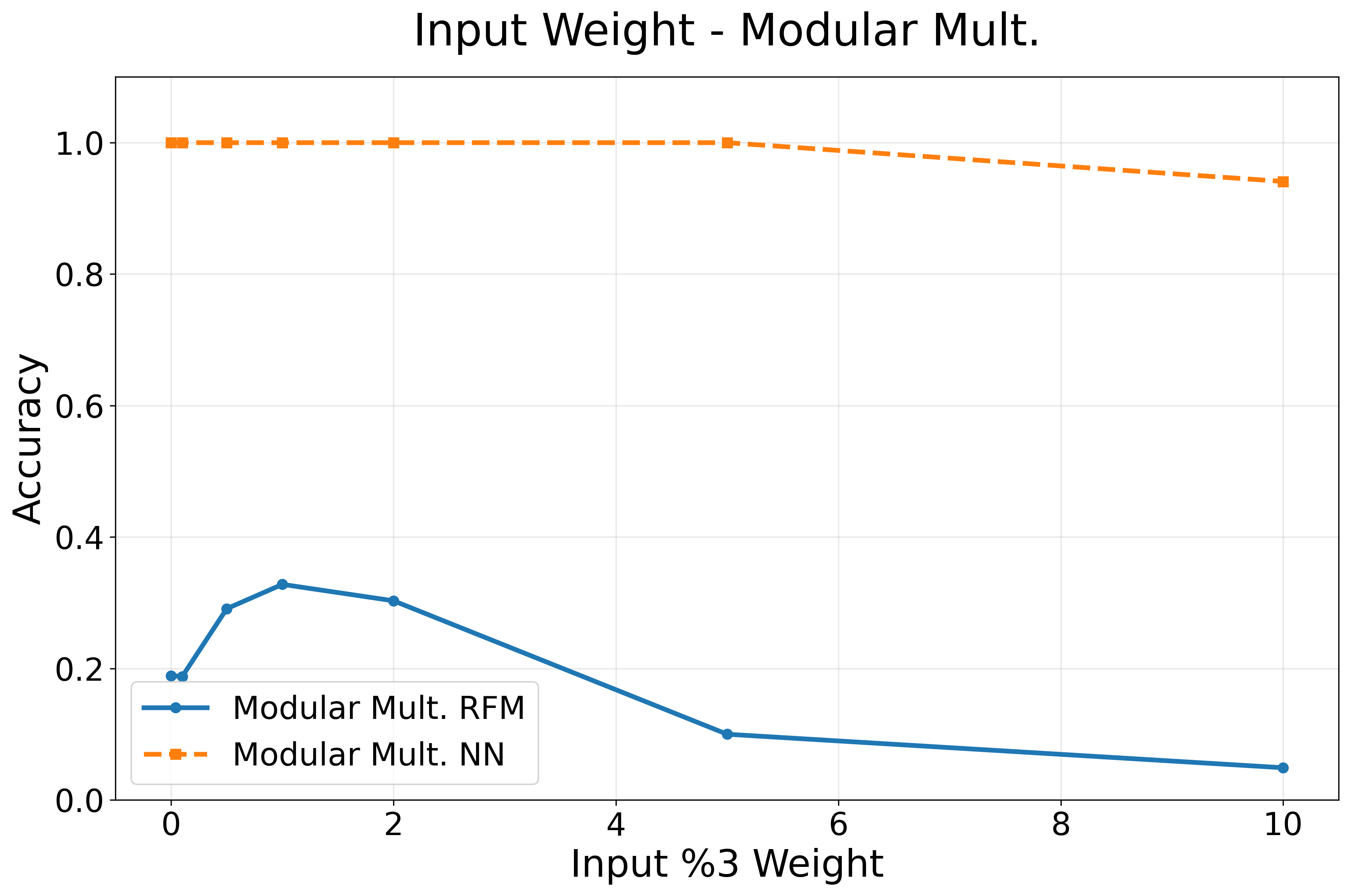}
    \end{subfigure}
    \begin{subfigure}{0.4\textwidth}
        \centering
        \includegraphics[width=\linewidth]{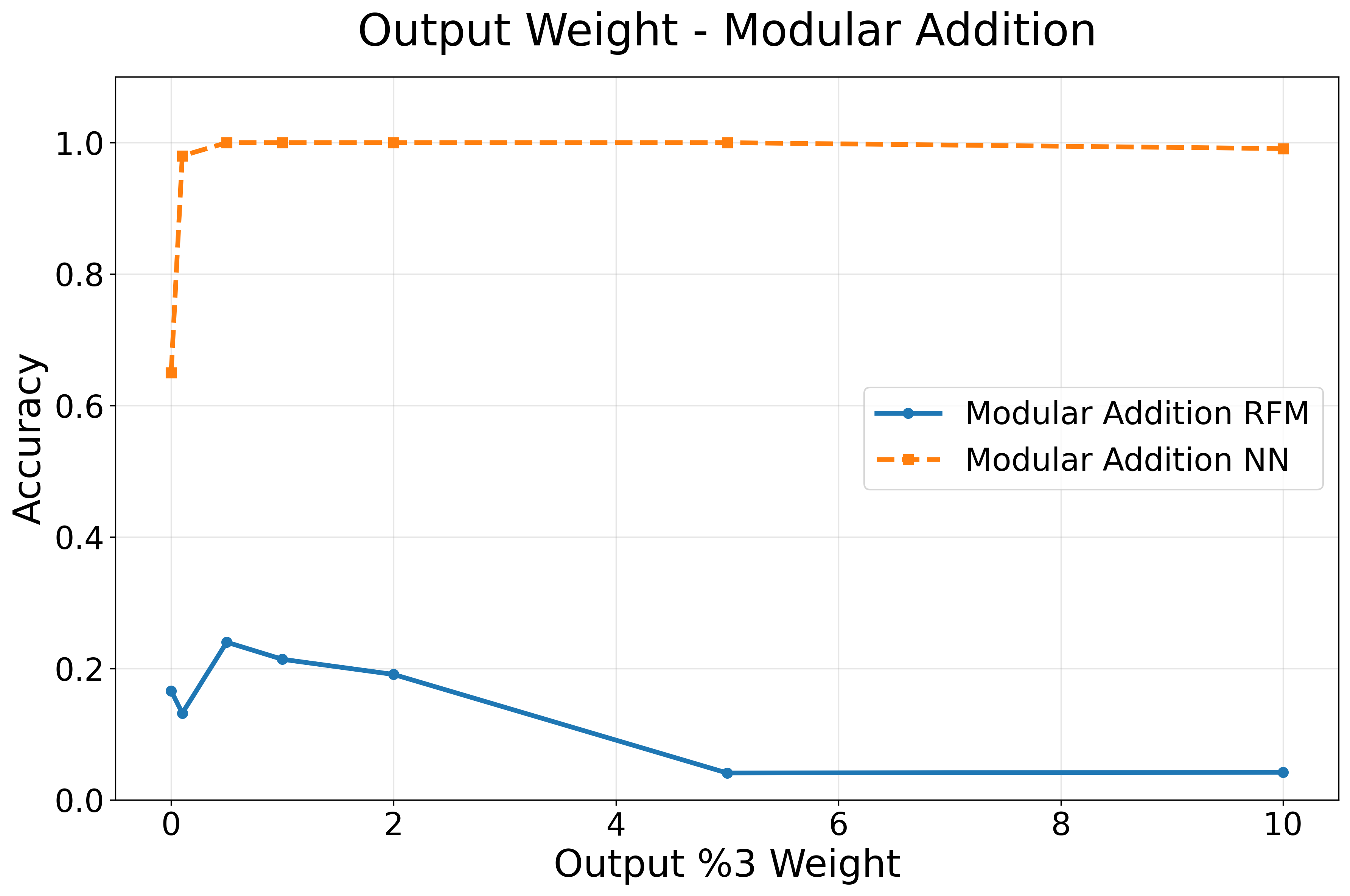}
    \end{subfigure}
    \begin{subfigure}{0.4\textwidth}
        \centering
        \includegraphics[width=\linewidth]{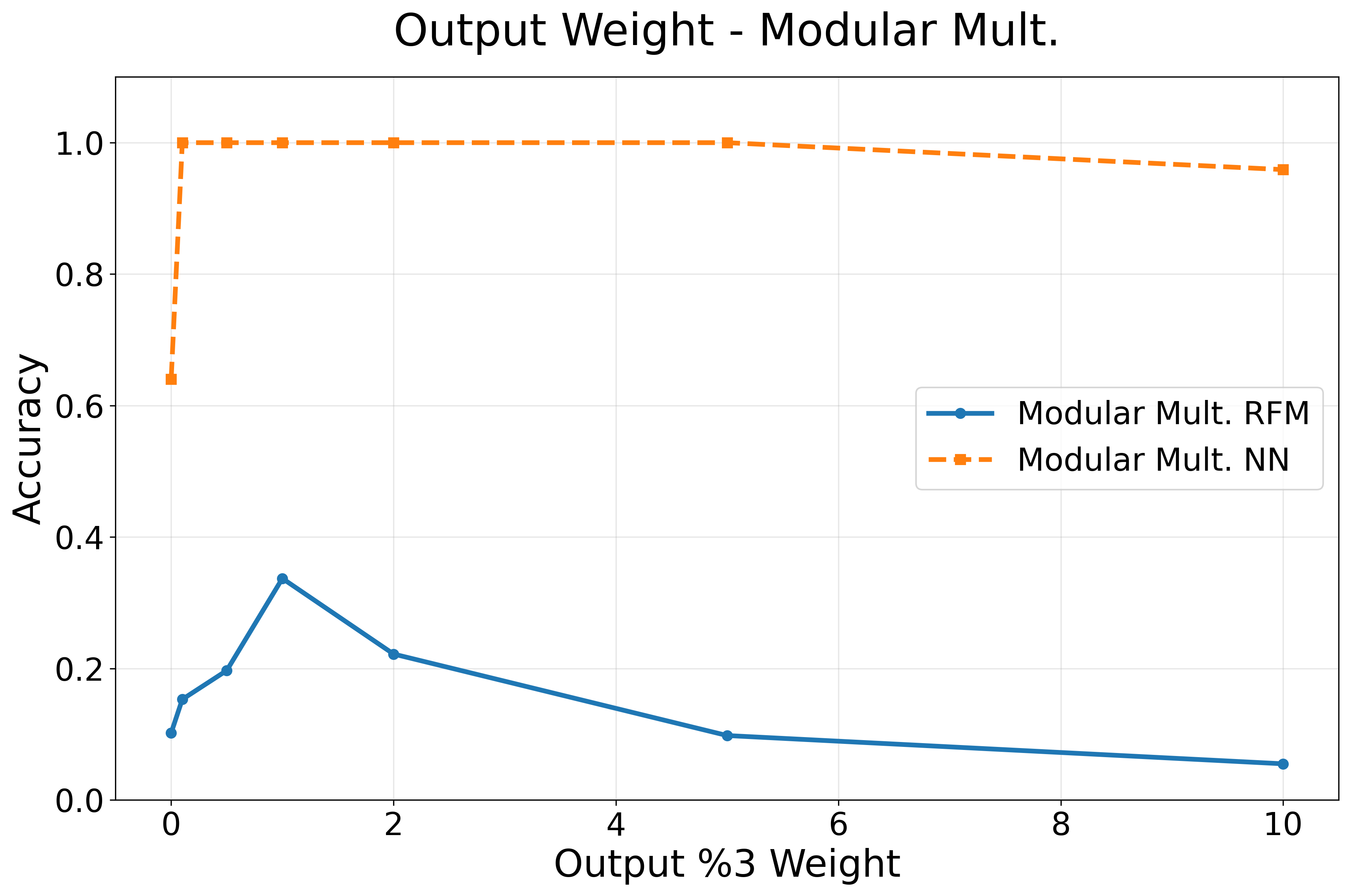}
    \end{subfigure}
    
    \caption[RFM vs. neural network performance on various reweightings of divisible-by-3 input pairs and labels...]{RFM vs. neural network performance on various reweightings of divisible-by-3 input pairs and labels. All ``unweighted" elements are assigned a weight of 1. \textbf{Top row}: RFM vs. NN on modular addition and multiplication (p=61) for various input weights of pairs divisible by 3. As can be observed, any reweighting greatly damages RFM performance. \textbf{Bottom Row}: RFM vs. NN on modular addition and multiplication (p=61) for various label weights of labels divisible by 3. While neural networks are impacted by outright exclusion, they are far less sensitive to non-uniformity/resampling.}
    \label{imbalanced-input-output}
\end{figure}

In our second experiment, we instead apply our condition to our output class distribution, varying the weight of our divisible-by-3 output classes and observing the effects on the performance of each model. We report our addition and multiplication results for each task in Figure \ref{imbalanced-input-output}.

Our figure elucidates several very interesting phenomena. Firstly, the figure shows that \textbf{any} resampling scheme, including a uniform one, greatly damages RFM results. The presence of some pairs/labels in higher multiplicity while less unique pairs appear overall causes the RFM - which achieves perfect training accuracy - to overfit on the present pairs. As the weight of pairs or labels divisible by 3 increases to 5 or 10 — meaning they become 5 or 10 times more likely than an average training sample to be selected — the RFM increasingly overfits to these examples, and test performance declines further. Neural networks, while impacted by the complete exclusion of a third of the output classes, are otherwise unaffected.

\begin{figure}
    \centering
    \begin{subfigure}{0.6\textwidth}
        \centering
        \includegraphics[width=\linewidth]{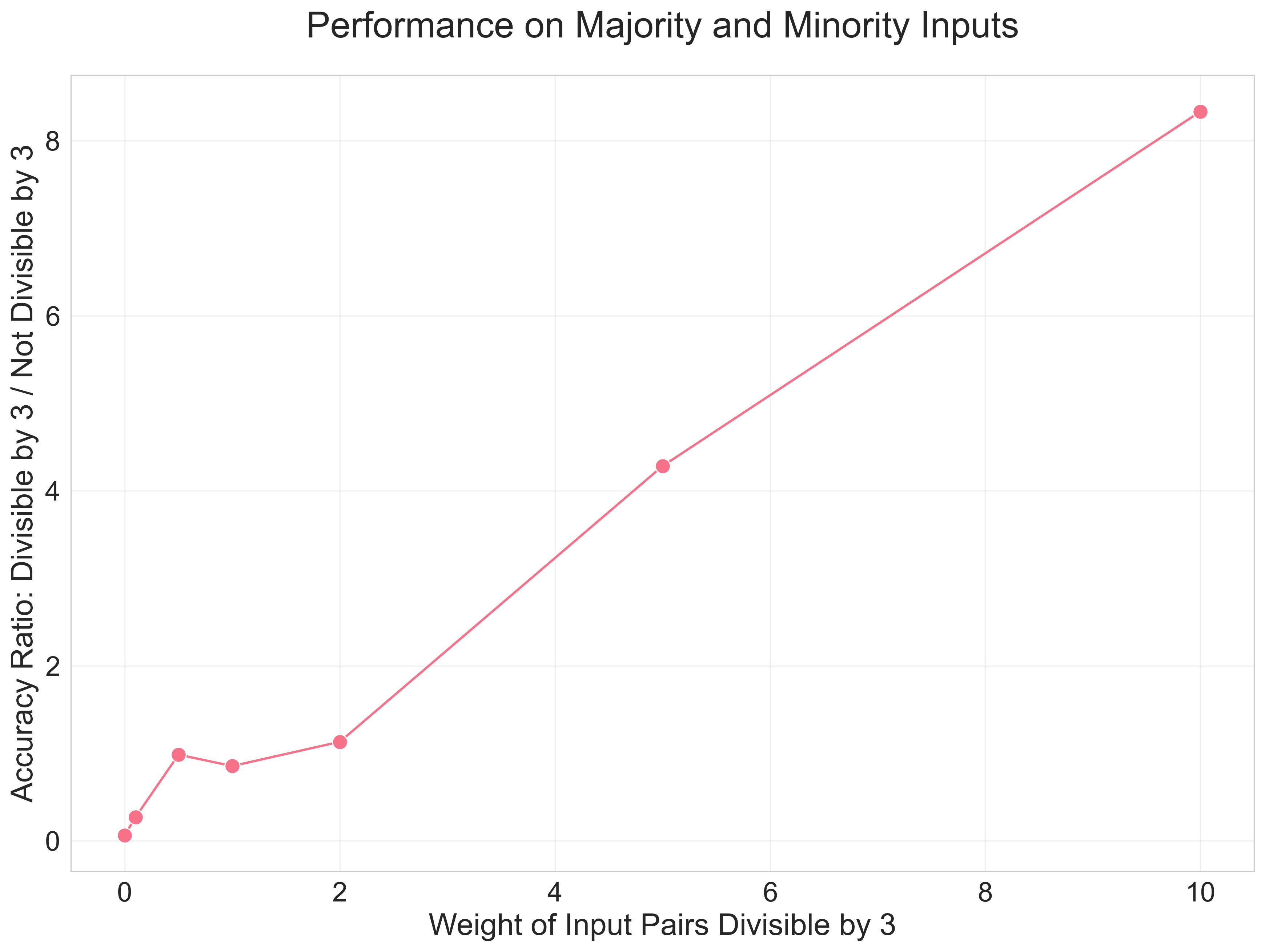}
    \end{subfigure} 
    \caption[Ratio of RFM modular addition performance on divisible-by-3 inputs vs. not-divisible-by-3 inputs as the weight of divisible-by-3 inputs in the training set increases...] {Ratio of RFM modular addition performance on divisible-by-3 inputs vs. not-divisible-by-3 inputs as the weight of divisible-by-3 inputs in the training set increases. As the weight becomes much larger, the vast majority of what the RFM learns becomes predicated on inputs divisible by 3.}
    \label{imbalanced-input-performance-ratio}
\end{figure}

We present evidence of the overfitting in Figure 
\ref{imbalanced-input-performance-ratio}, where it is shown that as the divisible-by-3 sample weight gets large, such samples begins to account for the vast majority of the performance of the RFM. We note that the absolute performance of both divisible-by-3 inputs and not-divisible-by-3 inputs tends downards as the weight of divisible-by-3 inputs increases, since the RFM begins to memorize its now-repetitive training data and loses generality.

\begin{figure}
    \centering
    \begin{subfigure}{0.6\textwidth}
        \centering
        \includegraphics[width=\linewidth]{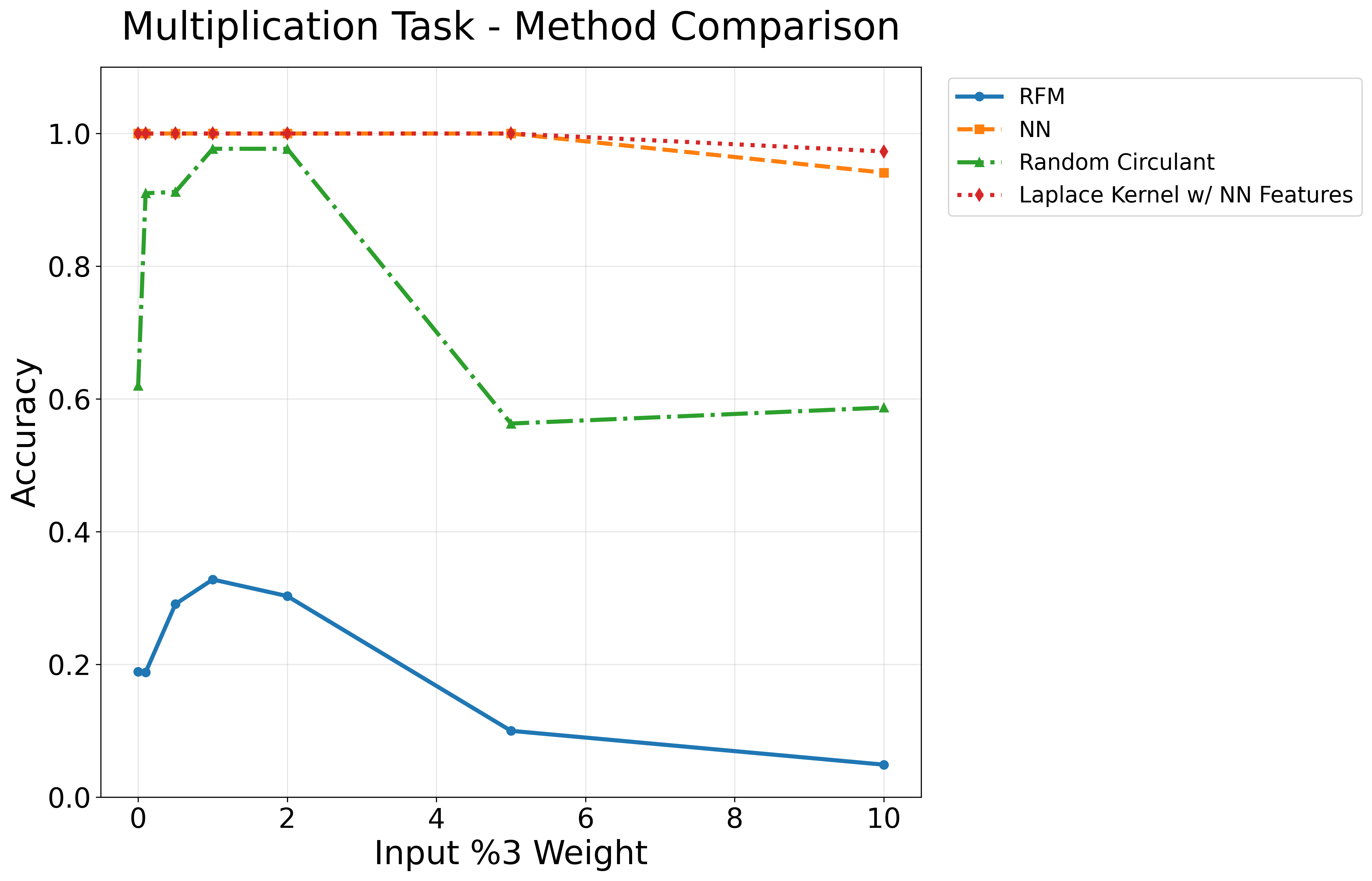}
    \end{subfigure}
    \caption[A comparison of methods on a modular multiplication task with p=61 and various weights for divisibile-by-3 inputs (input sampling with replacement)...]{A comparison of methods on a modular multiplication task with p=61 and various weights for divisibile-by-3 inputs (input sampling with replacement). Simply incorporating random circulant features bridges a significant portion fo the gap, and training on first layer post-nonlinearity features exceeds neural network performance altogether.}
    \label{rand-circulant-nn-features-unbalanced}
\end{figure}

Similarly to our previous section, we also conduct a test as to whether these phenomena can be accounted for by the feature learning capabilities of the first layer or simply inserting random circulant structure into the feature space. We observe - in similar fashion to our label noise experiments - that the gap is indeed one of feature learning in the neural network's first layer. A Laplace kernel, trained on nonlinearities of first layer features from the neural network, was able to match or surpass the performance of the neural network as a whole at every divisible-by-3 input ratio. Further, simply inserting random circulant structure into our features bridges a significant portion of the gap between RFMs and neural networks. We report both results in Figure \ref{rand-circulant-nn-features-unbalanced}.

.

\subsection{Systematic Exclusion}
\begin{figure}[h]
    \centering
    \begin{subfigure}{0.6\textwidth}
        \centering
        \includegraphics[width=\linewidth]{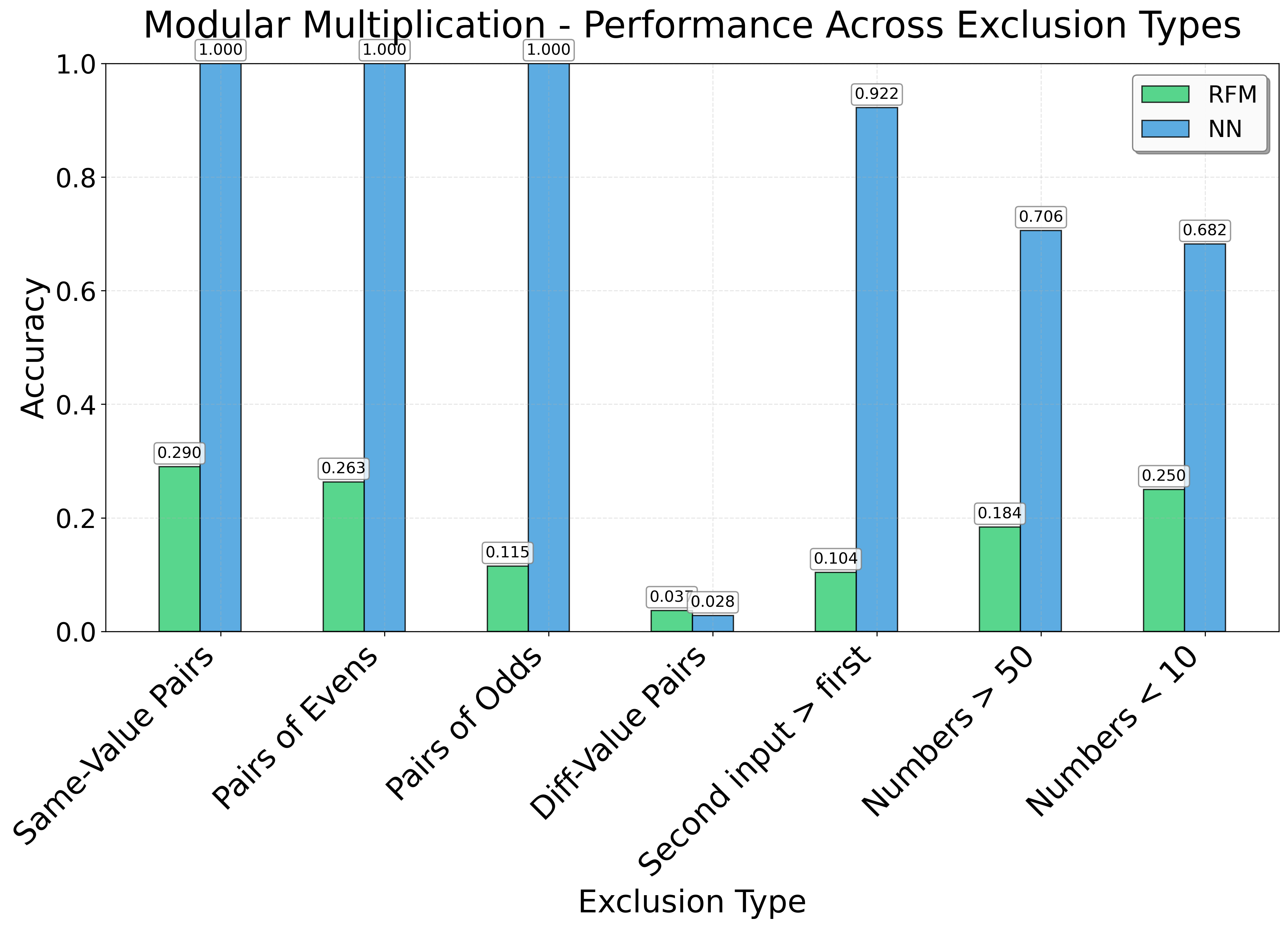}
    \end{subfigure}
    \caption[A comparison of methods on a modular multiplication task with p=61 on various forms of input exclusion...]{A comparison of methods on a modular multiplication task with p=61 on various forms of input exclusion. The neural network handily beats the RFM in 6/7 tasks, and struggles only when input pairs with different operands are excluded.}
    \label{exclusions-results}
\end{figure}

While the imbalancing of inputs and outputs divisible by 3 sheds signficiant light on the disparity between Neural Networds and RFMs, it is a somewhat random form of imbalance. To see whether different forms of imbalances would behave differently, we opted to experiment with seven forms of systematic exclusions: Exclusion of pairs where operands had the same value (e.g. (2,2)), exclusion of pairs where operands had different values, exclusion of odd/even numbers, exclusion of pairs where the second operand was greater than the first, exclusion of operands $<$10, and exclusion of operands $>$ 50. For each exclusion, all other data was given uniform weight and the RFM and neural network were evaluated across modular arithmetic tasks and GCD. The results strongly corroborate our other findings and show a signficant gap between RFMs and neural nets when dealing with data imbalanced mathematical tasks.  We report a detailed comparison in Figure \ref{exclusions-results}.

.

\section{Alternative Data Representation}
Having tested the label noise and data balance as two axes of data modification, we provide a final test for the RFM in the form of modifying our input representation. Instead of encoding our input operands as one-hot vectors between 0 and p, we create a more efficient representation using the Chinese Remainder Theorem to simulate a compute constrained scenario. We'll denote the Chinese Remainder Theorem as CRT from this point on.

The CRT claims that if we have a set of n integers $\{m_1, m_2, ...m_n\}$ such that $gcd(m_i, m_j)=1 \space$ $\forall i,j \in [n]$ (pairwise coprime), then all numbers in [0,$(\prod_{i=1}^{n}m_i)-1$] can be uniquely represented by the tuple (x mod $m_1$, x mod $m_2$,..., x mod $m_n$). Using a prime of 61, we leverage this property to encode our inputs as the tuple (x mod $3$, x mod $5$,..., x mod $7$), since all numbers in [0, (3*5*7)-1] = [0,104] will be uniquely encoded. Given we must encode two input operands, this allows us to reduce our representation size from 122 to 30 in theory, and from 122 to 42 in practice (we encode each modulo in a 7-dimensional vector for ease of computation).

This encoding has two interesting properties: it is both more efficient and increases the difficulty of the task, because the model might now need decode inputs from CRT to a normal numerical scale. It provides preliminary insights as to how these models might deal with more "densely" encoded mathematical data.

\begin{figure}[h]
    \centering
    \begin{subfigure}{0.71\textwidth}
        \centering
        \includegraphics[width=\linewidth]{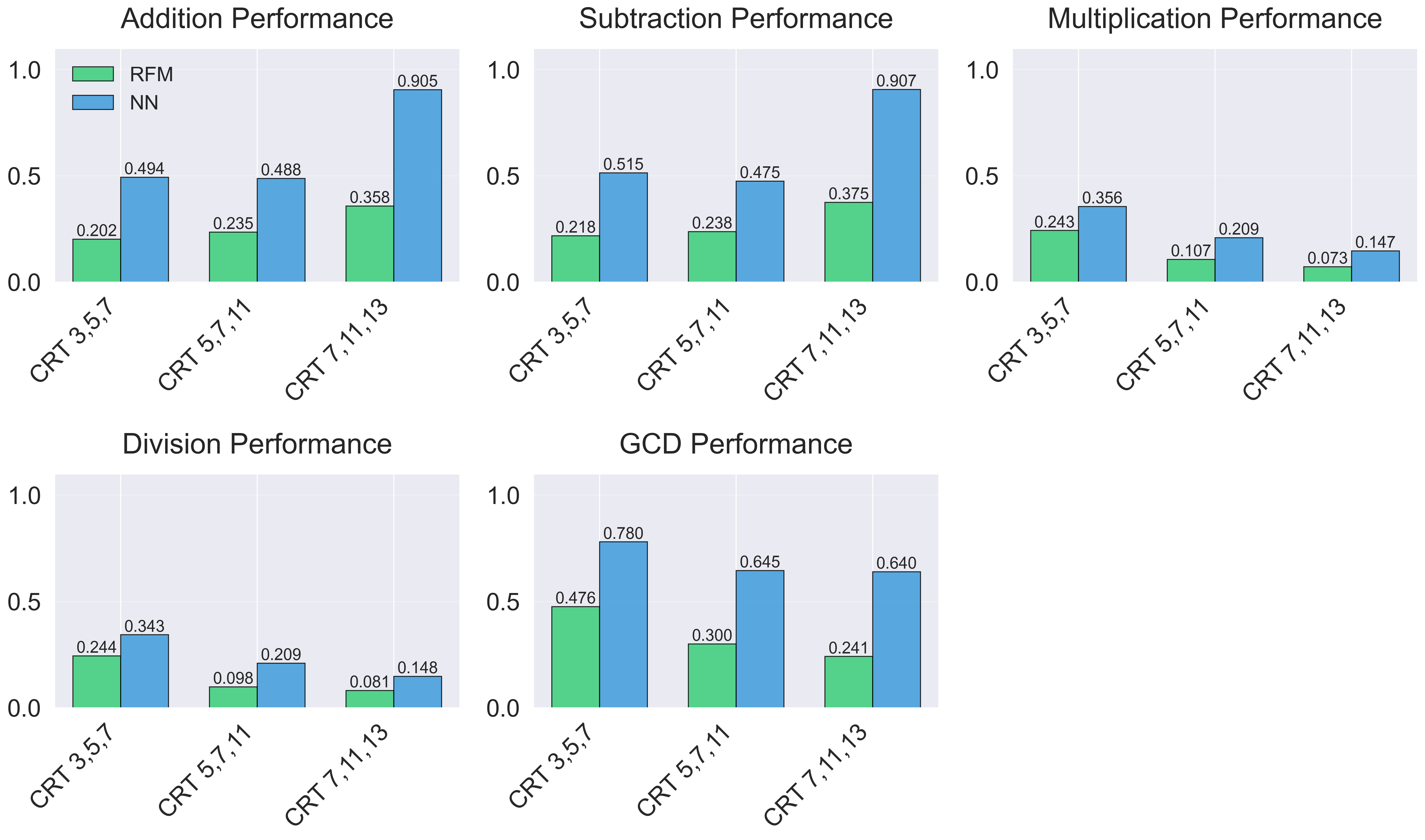}
    \end{subfigure}
    \begin{subfigure}{0.4\textwidth}
        \centering
        \includegraphics[width=\linewidth]{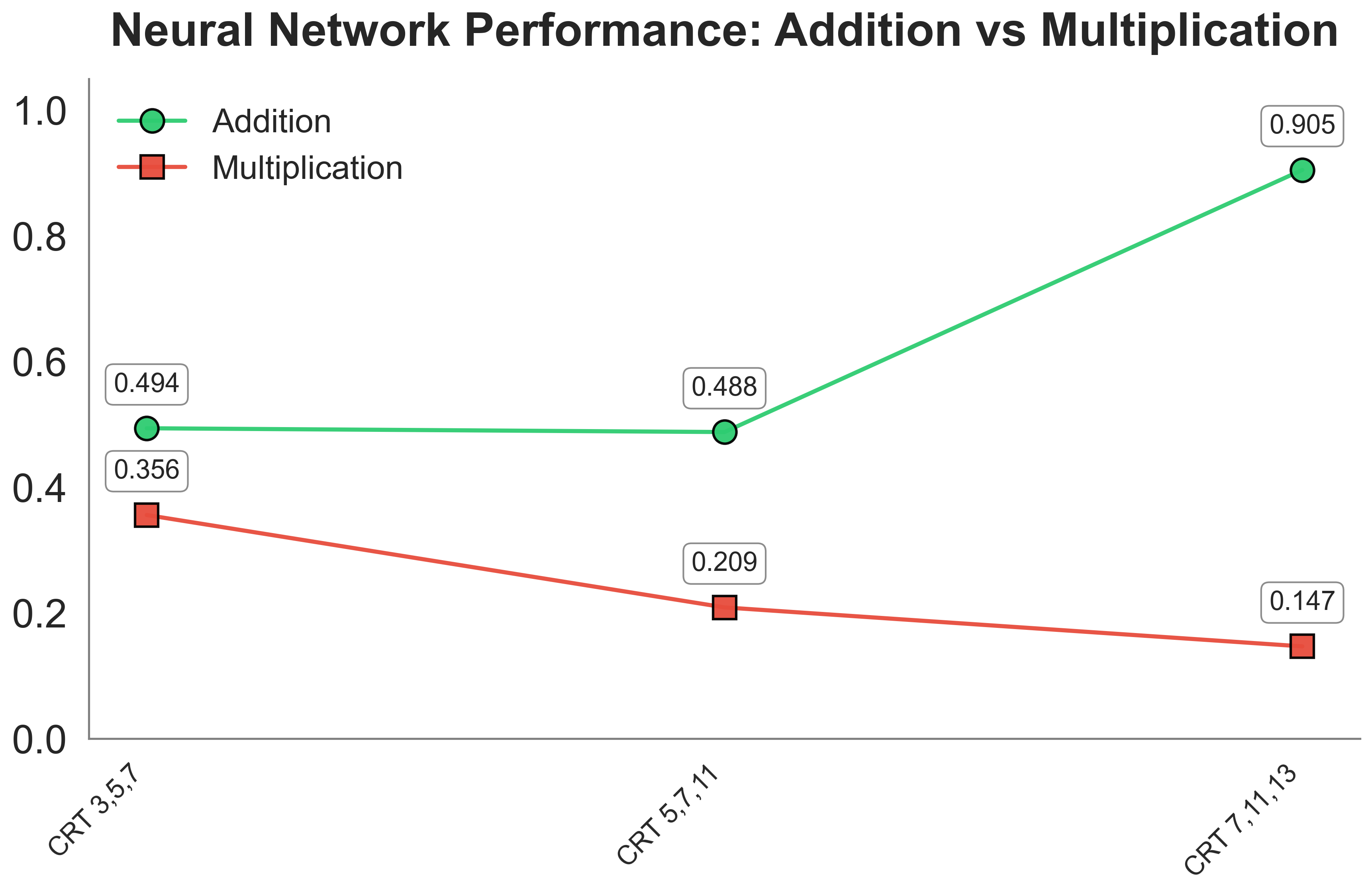}
    \end{subfigure}
    \begin{subfigure}{0.4\textwidth}
        \centering
        \includegraphics[width=\linewidth]{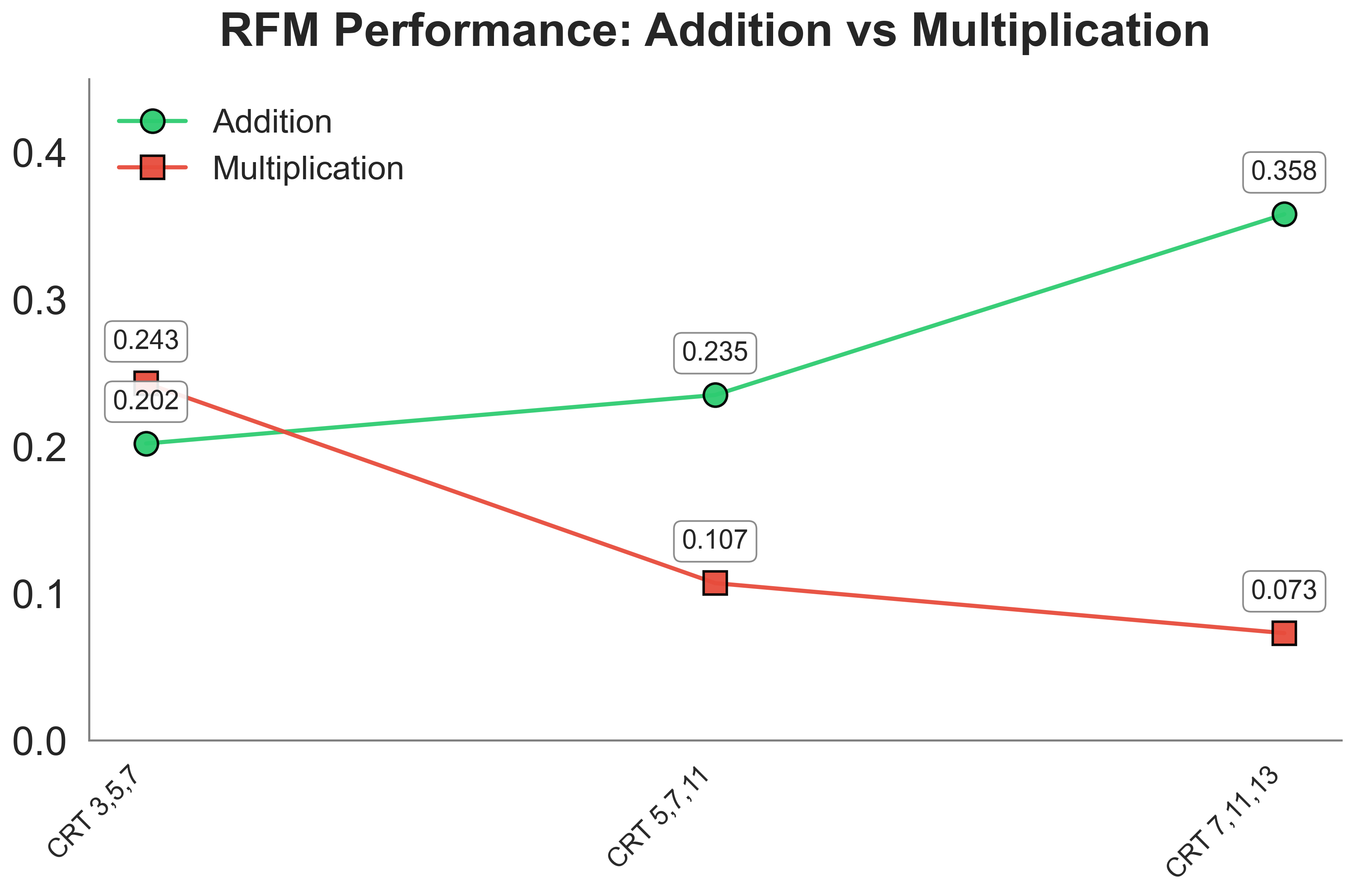}
    \end{subfigure}
    \caption[RFM vs. neural network performance on various CRT encodings...]{RFM vs. neural network performance on various CRT encodings. \textbf{Top}: A thorough performance comparison of neural networks and RFMs across all tasks. \textbf{Bottom}: A comparison of neural networks and RFMs across different CRT encodings. The results show that modular addition becomes an easier task for both models as the size of the CRT representation increases, whereas modular multiplication becomes more difficult.}
    \label{crt-results}
\end{figure}

We report results across all tasks and a side by side comparison of RFMs and neural networks in Figure \ref{crt-results}. To ensure that results are not unique to a single CRT representation, we utilize three forms, leveraging (3,5,7), (5,7,11), and (7,11,13) as our three triplets of coprimes.  

Interestingly, while neural networks certainly outperform RFMs for this representation style (particularly for larger representations on addition/subtraction), for several of the tasks the gap isn't very large - both models struggle mightily. Crucially, both models learn minimially and achieve their best performance in the first or second iteration/epoch.

Modular multiplication and division, in particular, are immensely difficult tasks for both models and worsen as the size of the CRT representation increases. This is to some extent a property of the encoding itself, as modular addition and subtraction can essentially be done elementwise with CRT representation, while multiplication / division may require large intermediates or decoding the representation altogether. Decoding, in particular, is a task that increases in difficulty as the size of the representation does, as there are more candidates for potential coprimes.

\begin{figure}[h]
    \centering
    \begin{subfigure}{0.6\textwidth}
        \centering
        \includegraphics[width=\linewidth]{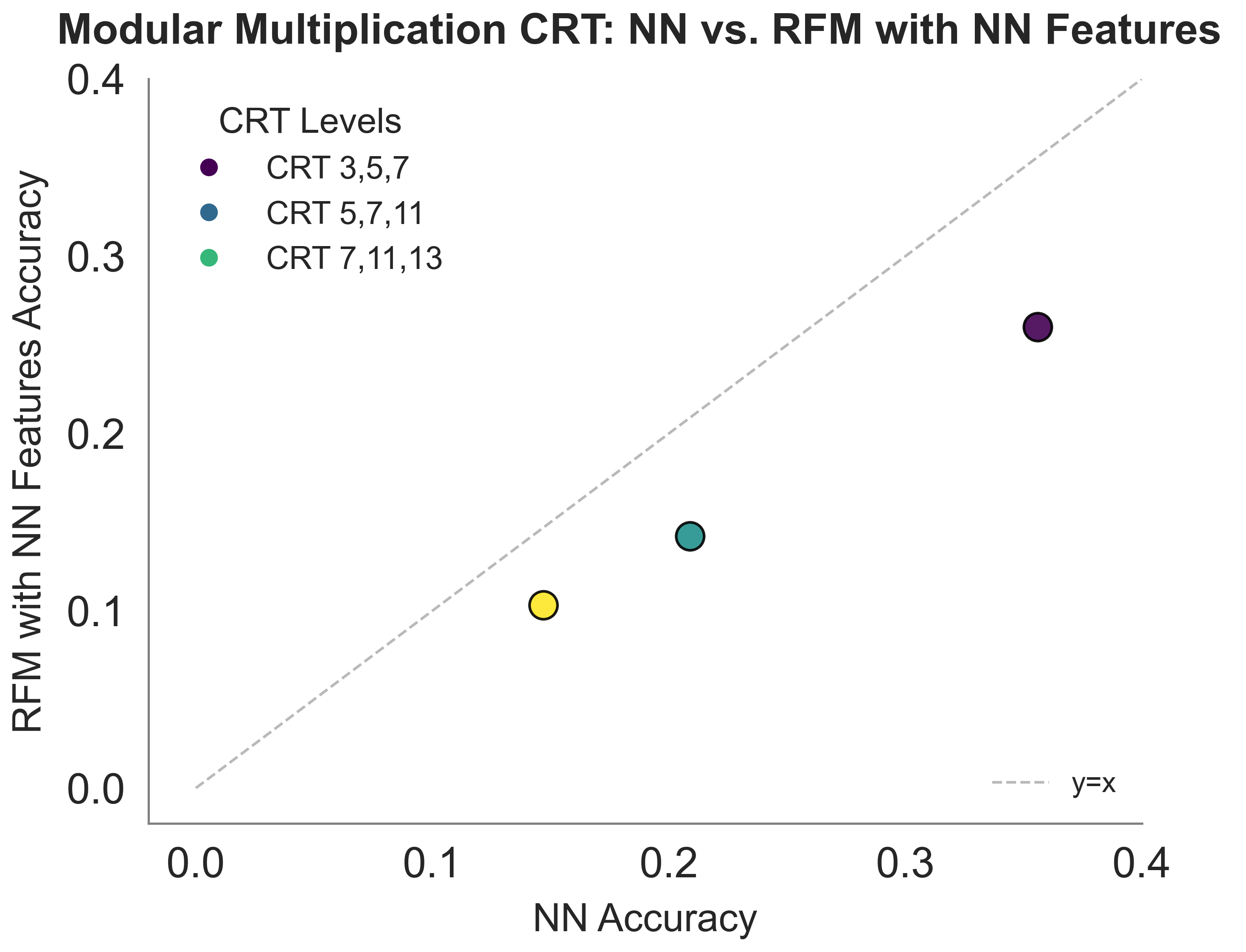}
    \end{subfigure}
    \caption[Laplace kernel with NN first Layer (post nonlinearity) features vs. neural network performance on various CRT encodings...]{Laplace kernel with NN first Layer (post nonlinearity) features vs. neural network performance on various CRT encodings. Both models are tasked with solving CRT-encoded modular multiplication with p=61. This time, the Laplace kernel using the first layer features underperforms the network itself, showing the gap isn't entirely one of first layer feature learning.}
    \label{CRT-Laplace-NN}
\end{figure}

As in the other sections, we check to see whether the gap is one of first layer feature learning and attempt to achieve top performance with first layer features and a Laplace kernel once again. Unlike previous contexts, this gap only seems partially attributable to first layer feature learning, as the kernel with first layer features still consistently underperforms the neural network. We provide results in figure \ref{CRT-Laplace-NN}.

%% file: 4_k_inverse.tex
\chapter{Bridging the Gap: The K-Inverse-RFM}
In the previous section, we showed the presence of a sizeable and important performance gap between RFMs and neural networks across a set of changes to the noise level, distribution, and structure of the data. Crucially, we demonstrated that a kernel trained on first-layer neural network features can consistently close—and sometimes exceed—the accuracy of the original network. This suggests a shortfall in the RFM’s native feature learning capacity. To address this, we present the \textbf{K-Inverse-RFM}.

We note that our version of \textbf{K-Inverse-RFM} is an evolution of a framework created by Mallinar \& Belkin \cite{k_inverse_rfm_original}. The original framework proposes a projection of the label onto the kernelized feature space with a power of -1, and a simple moving average of AGOP matrices for improved stability.  While this framework often yields improvements in many multiclass scenarios, it still suffers from instability and does not generalize to some gaps introduced earlier. We introduce our own framework below.

\section{Method}
Multiclass feature learning is a known shortcoming of kernel methods. Kernels are fixed transformations on the input (they are not learned, but rather pre-applied), and kernel regression learns a separate classifier using the transformed inputs for every class. This has two implications. Firstly, if the data for one class is corrupted or learned improperly, we cannot hope to use features from other classes as a means of compensation. Secondly, a kernel method could necessitate the presence of a significant amount of data from \textbf{each} class for strong performance.

While the AGOP is a feature learning mechanism useful for amplifying certain relevant features in the dataset, it still only learns a linear re-weighting of the input. In order to best optimize the quality of our model, we should deal with its nonlinear component. We do this by modifying \ref{RFM} as follows:

\begin{align}
    y_p &= K(X, X; M_t)^{d}y              && (\textbf{Label Projection}) \\
    \alpha_t &=  (K(X, X; M_t)+\lambda I)^{-1}y_p        && \textbf{(Predictor Training)} \label{Predictor_Training} \\
    f_t(x) &= K(x, X; M_t) \alpha_t        && \\
    M_{t+1} &= \frac{\Sigma_{i=t-k}^{t}\space \frac{1}{l_i}G(f_t)^s}{\Sigma_{i=t-k}^{t}\space \frac{1}{l_{i}}}                  && \textbf{(Feature Learning)} \label{K-Inv-RFM}
\end{align}

Crucially, the K-Inverse methodology adds in three primary changes. Firstly, we now solve a new equation of the form: $$\alpha(K(X, X; M_t)+\lambda I) = K(X, X; M_t)^dy$$ Setting $d=-1$ yields the name K-Inverse, and also reveals the nature of the formulation: we're mapping the labels to the space of the kernelized features, so as to represent each label as a combination of the post-kernel data points. What this effectively does is to represent outputs as combinations of data points, meaning our new, more efficient mapping is between one combination of data points and another. This also means that our kernelized representations are now shared features that can be recombined with a simple reweighting to produce our new labels, promoting multiclass structure.

The other two modifications are practical: we find, perhaps as a result of predicting in a high dimensional feature space, that the method tends to yield instability and large gradient changes in the AGOP. To address this, we formulate our AGOP features as a moving average, weighted by inverse training loss so as to reduce the impact of bad updates. Further, we reformulate our kernelized regression as kernelized ridge regression, using a high ridge term to constrain the size of our updates and introduce training stability. Finally, we vary the value of $d$ depending on the use case: we find that for denser feature and label spaces, using smaller or positive powers of $d$ can help improve performance and maintain feature stability. For our CRT experiments, for instance, we employ $d=0.4$.

All modifications are efficient and do not significantly reduce the speed of training the K-Inverse-RFM. Just like an RFM, the K-Inverse-RFM can be efficiently trained on a CPU.

\section{Results}
We proceed to test our method against the standard RFM and neural networks across all three data corrupted scenarios. In each section, we provide result comparison figures across standard RFMs, K-Inverse-RFMs, and neural networks. We leverage the best performing RFM kernel (Gaussian for Label Noise, Quadratic otherwise) for our RFM ($\lambda$ = 0.00001) and K-Inverse-RFM ($\lambda$ = 0.01), and use a quadratic activation function for our neural network (h=512). All models are trained to convergence on 50\% of unique input pairs unless the data is resampled. At this point, the highest achieved test set performance is reported.
\subsection{Label Noise}
\begin{figure}[h]
    \centering
    \begin{subfigure}{0.4\textwidth}
        \centering
        \includegraphics[width=\linewidth]{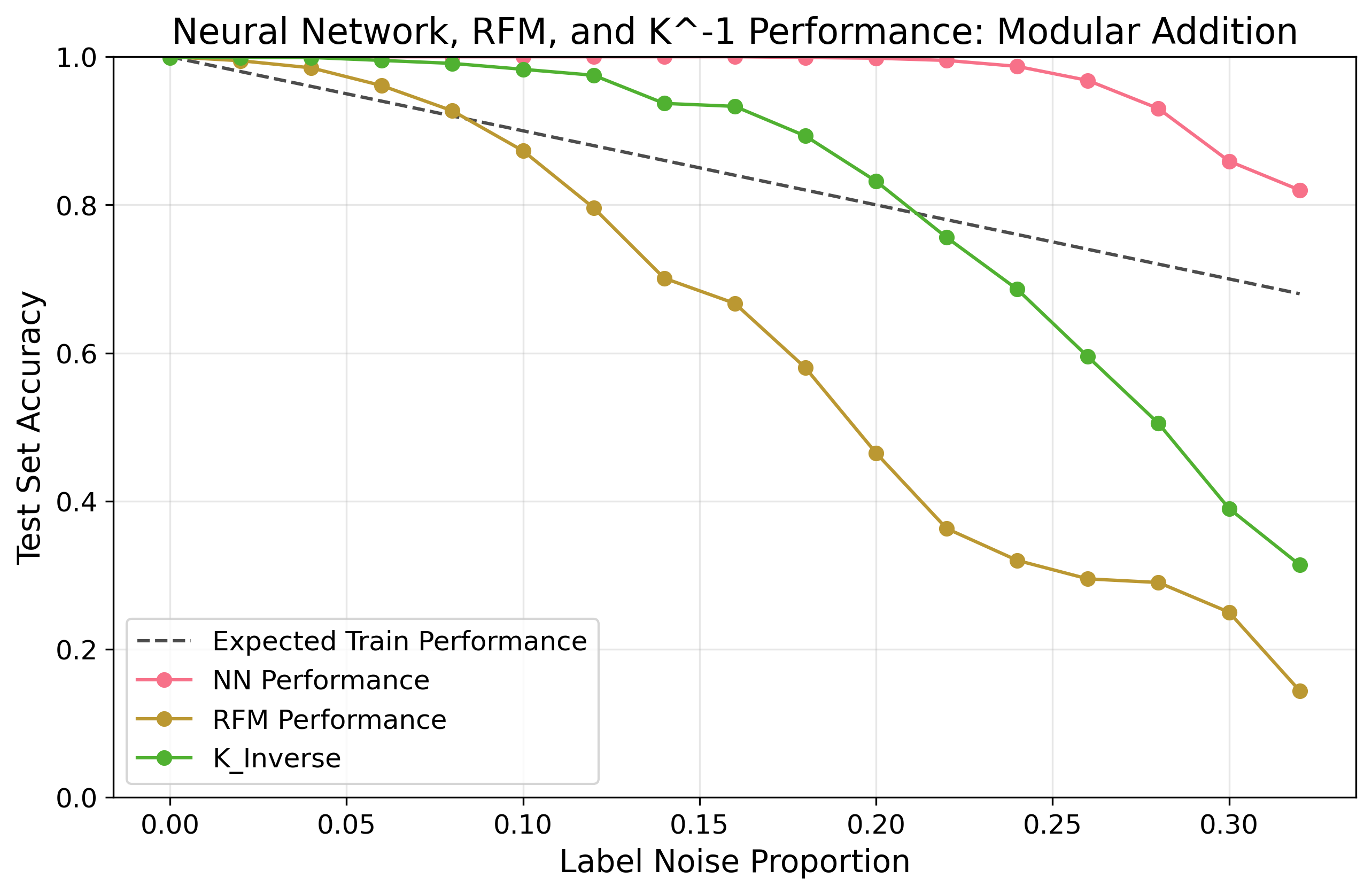}
    \end{subfigure}
    \begin{subfigure}{0.4\textwidth}
        \centering
        \includegraphics[width=\linewidth]{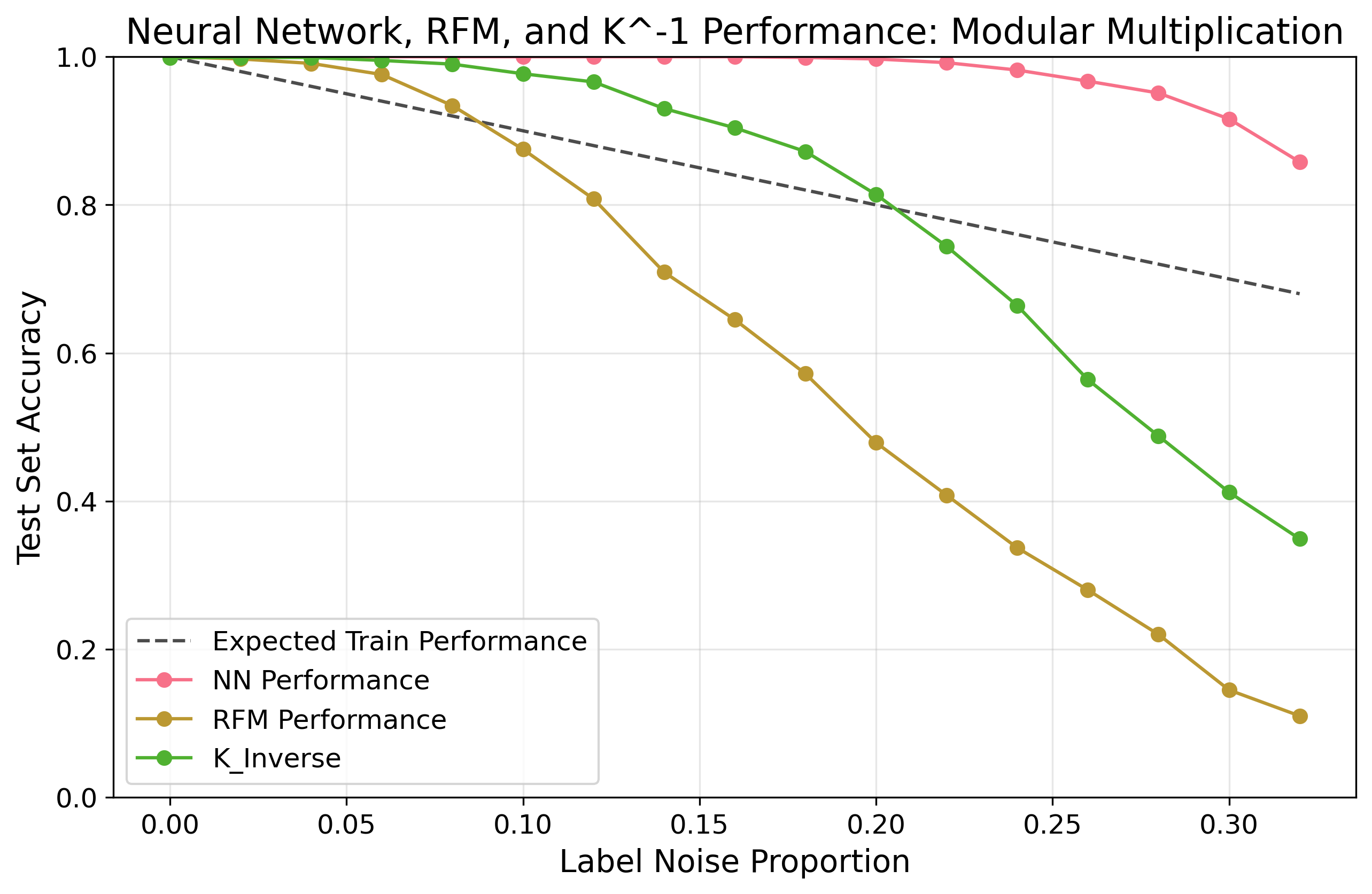}
    \end{subfigure}
    \caption[RFM, K-Inverse-RFM, and neural network performance comparison across modular addition and multiplication...]{RFM, K-Inverse-RFM, and neural network performance comparison across modular addition and multiplication. While it does not reach the performance of neural networks, the K-Inverse-RFM showcases much improved scaling relative to the RFM and on average bridges 64\% of the gap to the performance of neural networks.}
    \label{ln-k-inverse}
\end{figure}

In Figure \ref{ln-k-inverse} we show comprehensive results on label noise modular addition and multiplication across label noise levels ranging from 0-32\%. The K-Inverse-RFM bridges an average of 64\% of the gap between RFMs and neural networks, exhibiting significantly improved accuracy. It surpasses the RFM by as much as 39.3\% test accuracy and remains comparable in performance to neural networks until 14\% label noise. 

Notably, the K-Inverse-RFM shows much improved generalization capabilities relative to the RFM. We approximate this by comparing our results to the expected training performance of a perfect classifier (given $k$ label noise proportion, $E($train\_performance$)$ = $1-k$). If we fare better, then we have evidence that the model is learning to place emphasis on the non-random portion of the training date and generalize. While the RFM itself shows very little evidence of ever truly generalizing, the K-Inverse-RFM shows strong evidence of generalizing at intermediate levels of labels noise.

\subsection{Imbalanced Data}
\begin{figure}[h]
    \centering
    \begin{subfigure}{0.75\textwidth}
        \centering
        \includegraphics[width=\linewidth]{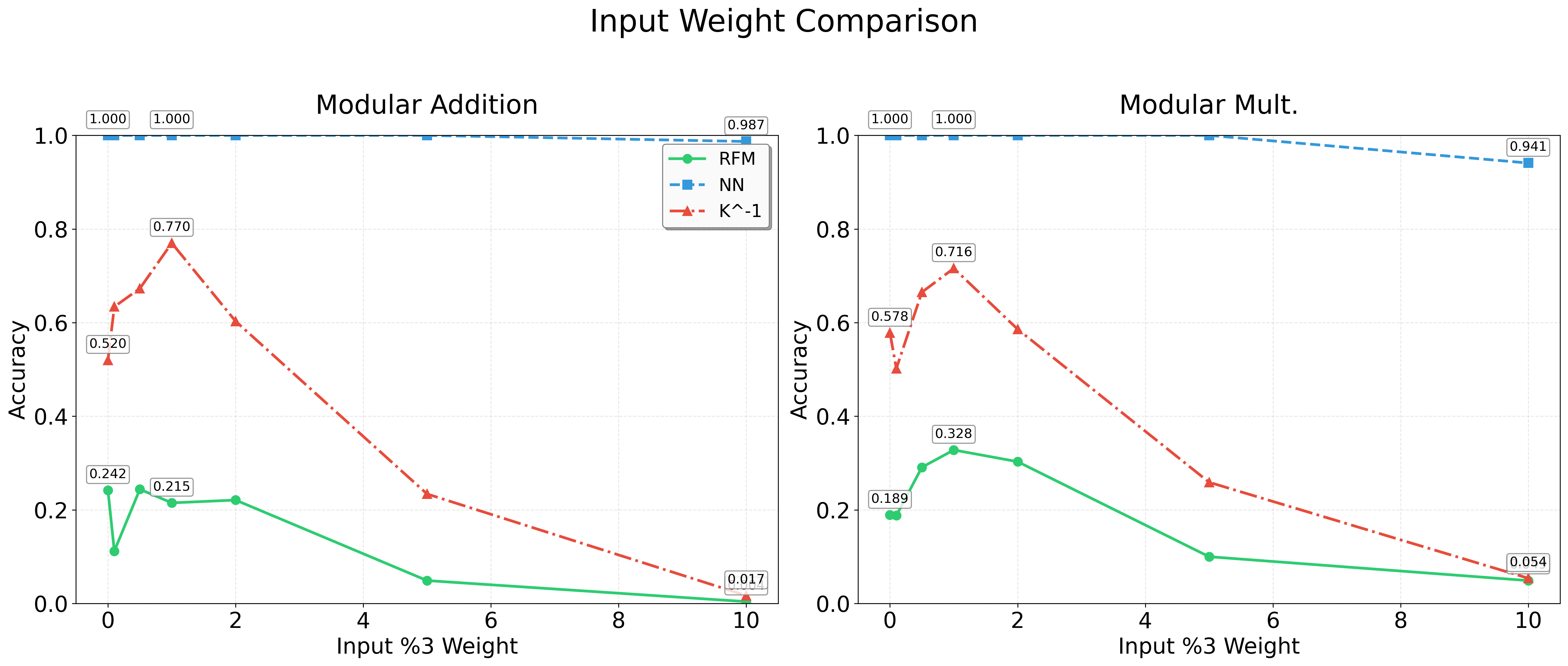}
    \end{subfigure}
    \begin{subfigure}{0.75\textwidth}
        \centering
        \includegraphics[width=\linewidth]{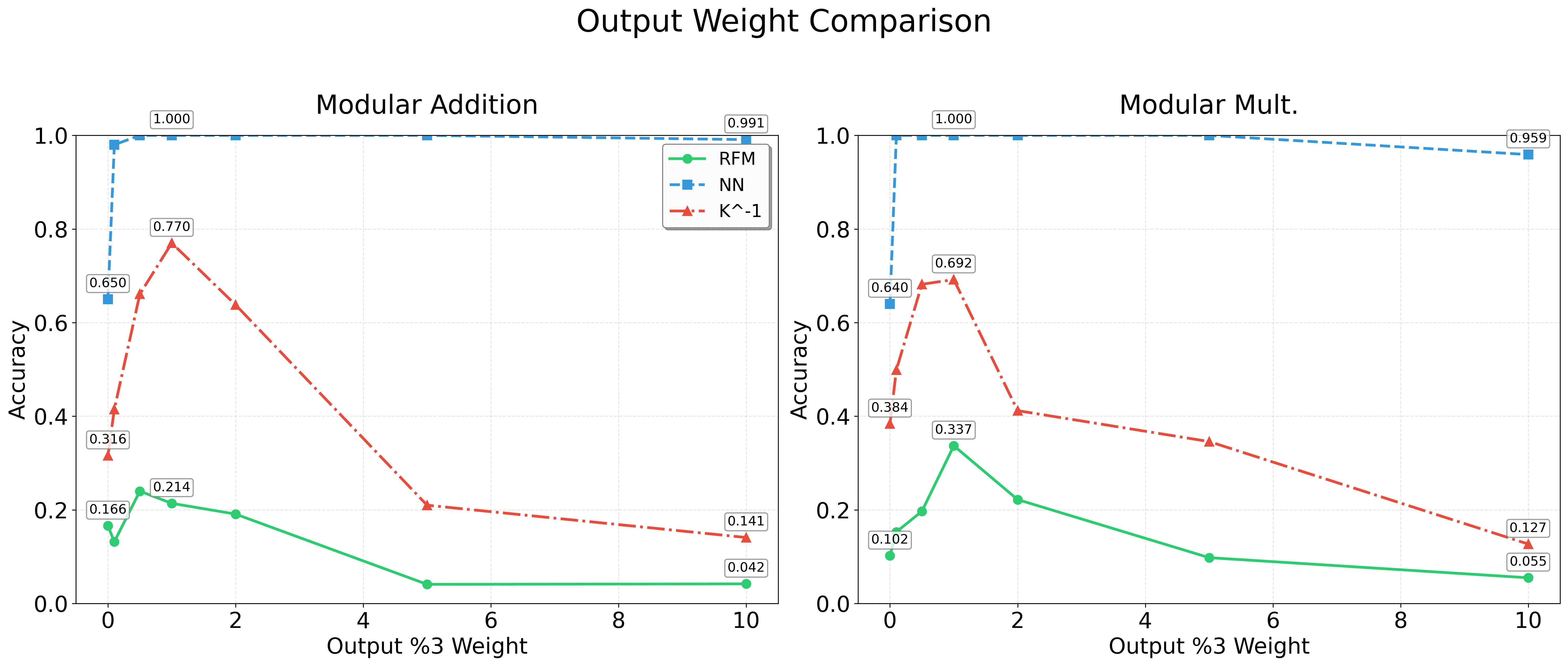}
    \end{subfigure}
    \caption[A comparison of RFM, K-Inverse-RFM and neural network performance for modular addition and multiplication...]{A comparison of RFM, K-Inverse-RFM and neural network performance for modular addition and multiplication. The top plot shows test set accuracy for different relative weightings of inputs divisible by 3, whereas the bottom plot shows test accuracy for different relative weightings of labels divisible by 3.}
    \label{k_inverse_unevens_div_3}
\end{figure}

Leveraging the same imbalances found in Section 3.1, we run the K-Inverse-RFM on various weights of divisible-by-3 inputs and labels. The story we uncover, shown in figure \ref{k_inverse_unevens_div_3}, is very much the same as the previous section: while the K-Inverse-RFM behaves like the RFM in its downward trend as data weight imbalances worsen, its actual performance is far better. At a divisible-by-3 input weight of 1 (which introduces uneveness into the data simply by sampling it with replacment), the improvement of K-Inverse-RFM over the standard RFM is a vast 53\%  jump in test accuracy.

\begin{figure}[h]
    \centering
    \begin{subfigure}{0.72\textwidth}
        \centering
        \includegraphics[width=\linewidth]{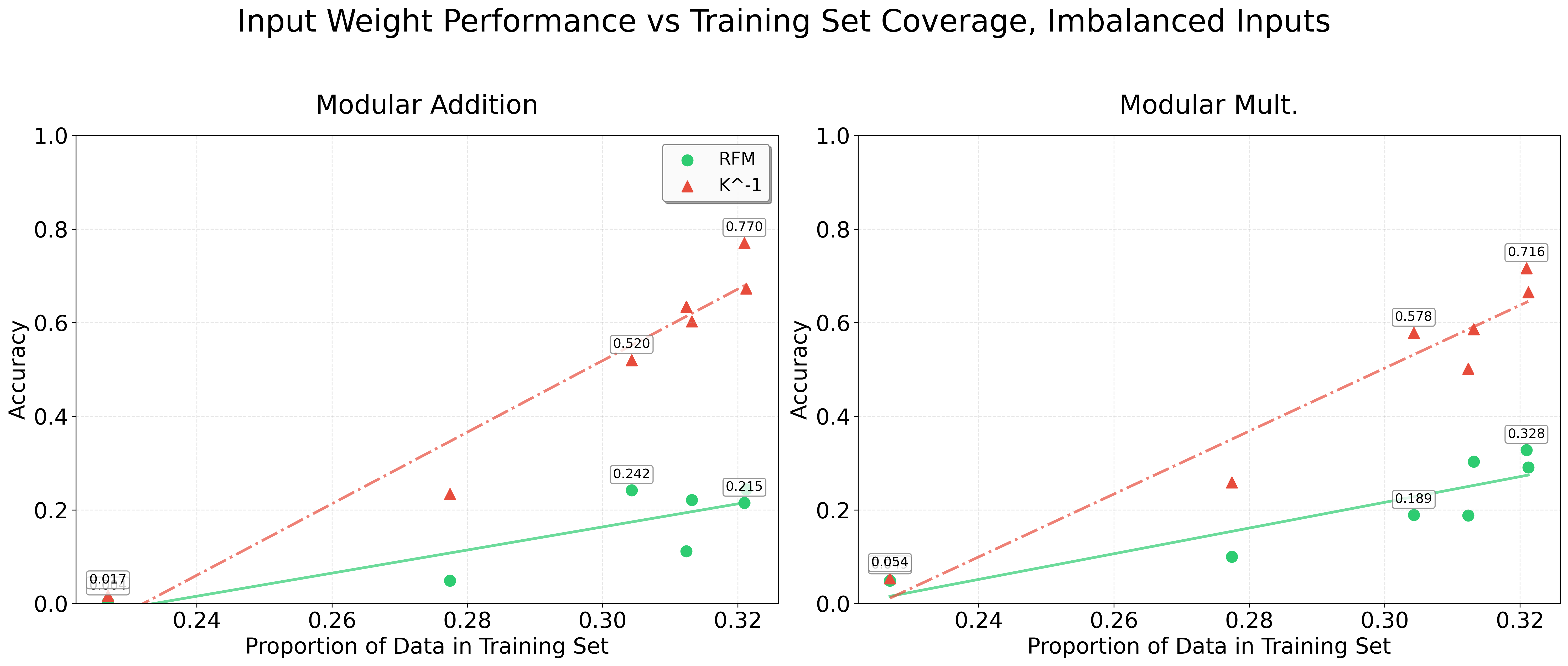}
    \end{subfigure}
    \begin{subfigure}{0.72\textwidth}
        \centering
        \includegraphics[width=\linewidth]{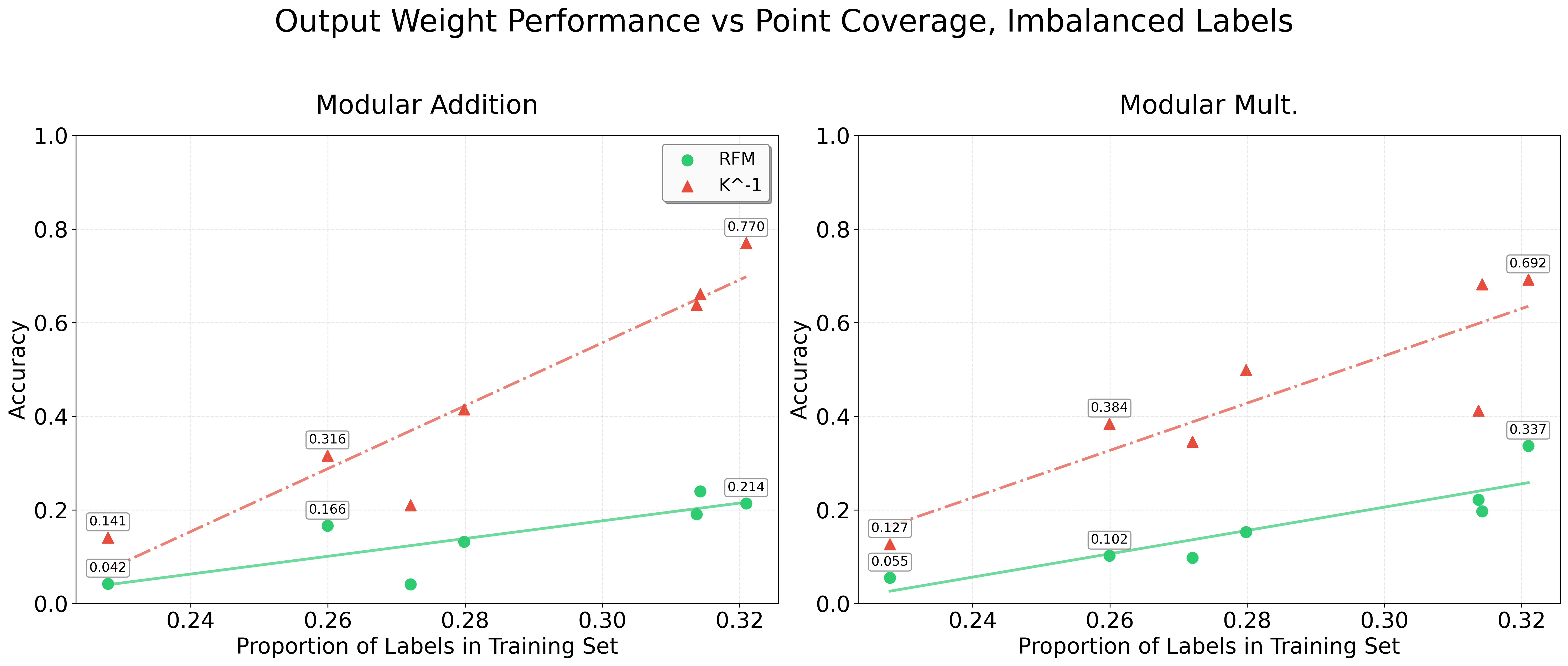}
    \end{subfigure}
    \caption[A reframing of Figure \ref{k_inverse_unevens_div_3} in terms of the number of unique points in each uneven sample...]{A reframing of Figure \ref{k_inverse_unevens_div_3} in terms of the number of unique points in each uneven sample. All plotted lines are lines of best fit. Both K-Inverse and RFM scale essentially linearly with the number of unique points in the sample, with the primary difference being that K-Inverse has a higher scaling factor (sample efficiency).}
    \label{k_inverse_unique_point_scaling}
\end{figure}

While training the RFM and K-Inverse-RFM, we also discovered that the repeated data points given by sampling with replacement make negligible to no difference in final RFM performance. Therefore, we removed these points altogether for faster training and ran follow-up experiments to ascertain whether the decay in RFM performance is a result of actual imbalances in the data or simply a lowered number of unique points. Our results, shown in \ref{k_inverse_unique_point_scaling}, surprisingly elucidate that performance in these contexts scales almost linearly with the number of unique training points: the actual distribution of the points matters less than the sheer number of unique samples. 

The results clarify why K-Inverse-RFM outperforms standard RFM. Although both models scale linearly with the number of training samples, K-Inverse-RFM achieves greater performance gains per sample. This indicates improved sample efficiency, which can be attributed to its ability to effectively share information across classes.

\begin{figure}[h]
    \centering
    \begin{subfigure}{0.6\textwidth}
        \centering
        \includegraphics[width=\linewidth]{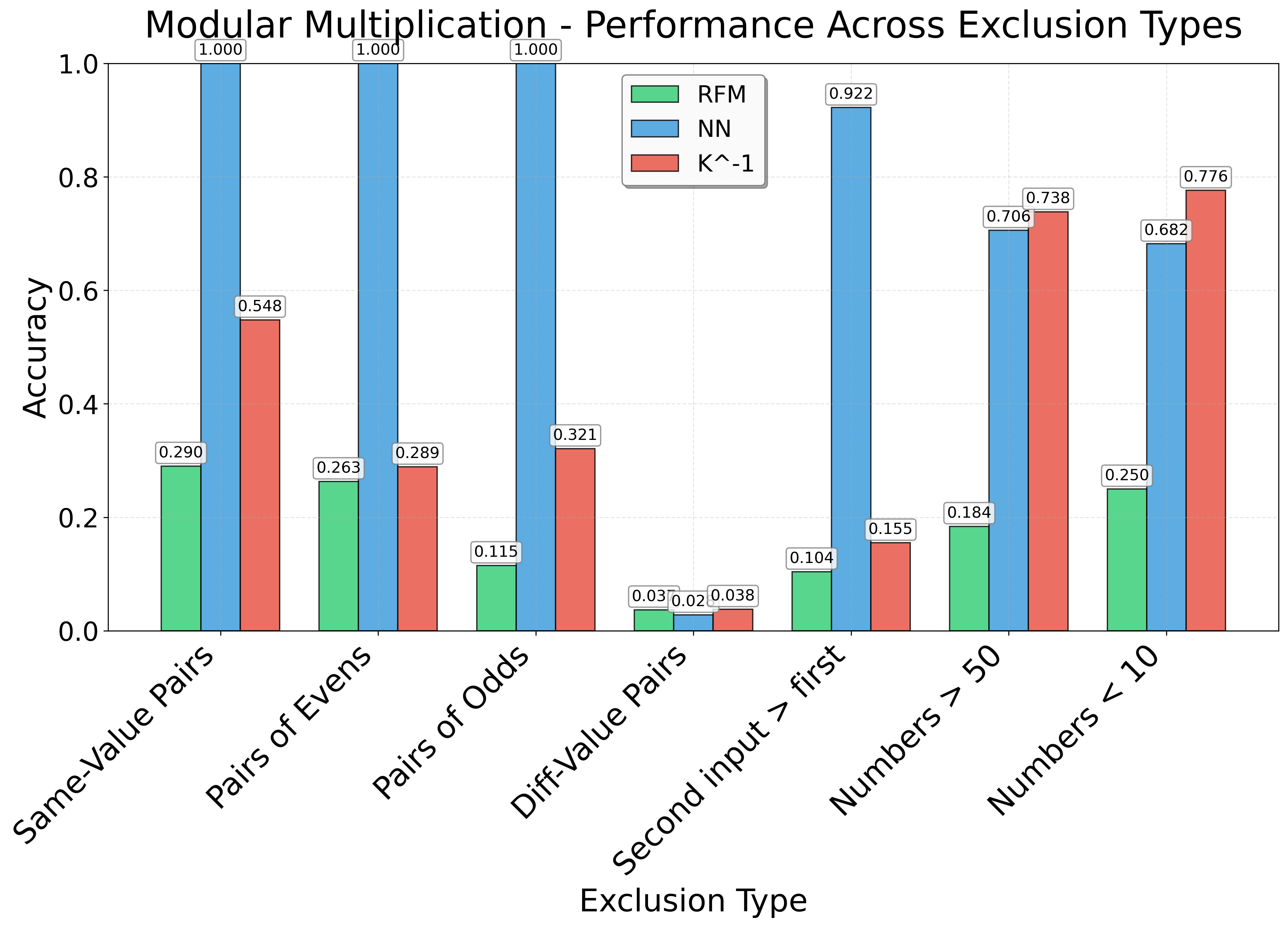}
    \end{subfigure}
    \caption[A comparison of K-Inverse, RFMs, and neural networks across different excluded inputs for modular addition...]{A comparison of K-Inverse, RFMs, and neural networks across different excluded inputs for modular addition. K-Inverse-RFMs outperform RFMs across the board, in particular on tasks with a larger number of unique input points.}
    \label{k_inverse_exclusions}
\end{figure}

We complete our imbalanced data comparisons by reviewing results across input exclusion types introduced in \ref{exclusions-results}. K-Inverse-RFMs once again beat RFMs across the board, with particularly impressive results coming on exclusions of Numbers $>$ 50 and Numbers $<$ 10. K-Inverse-RFM exclusion results are given in figure
\ref{k_inverse_exclusions}.

\subsection{Alternative Data Representation}
\begin{figure}[h]
    \centering
    \begin{subfigure}{0.68\textwidth}
        \centering
        \includegraphics[width=\linewidth]{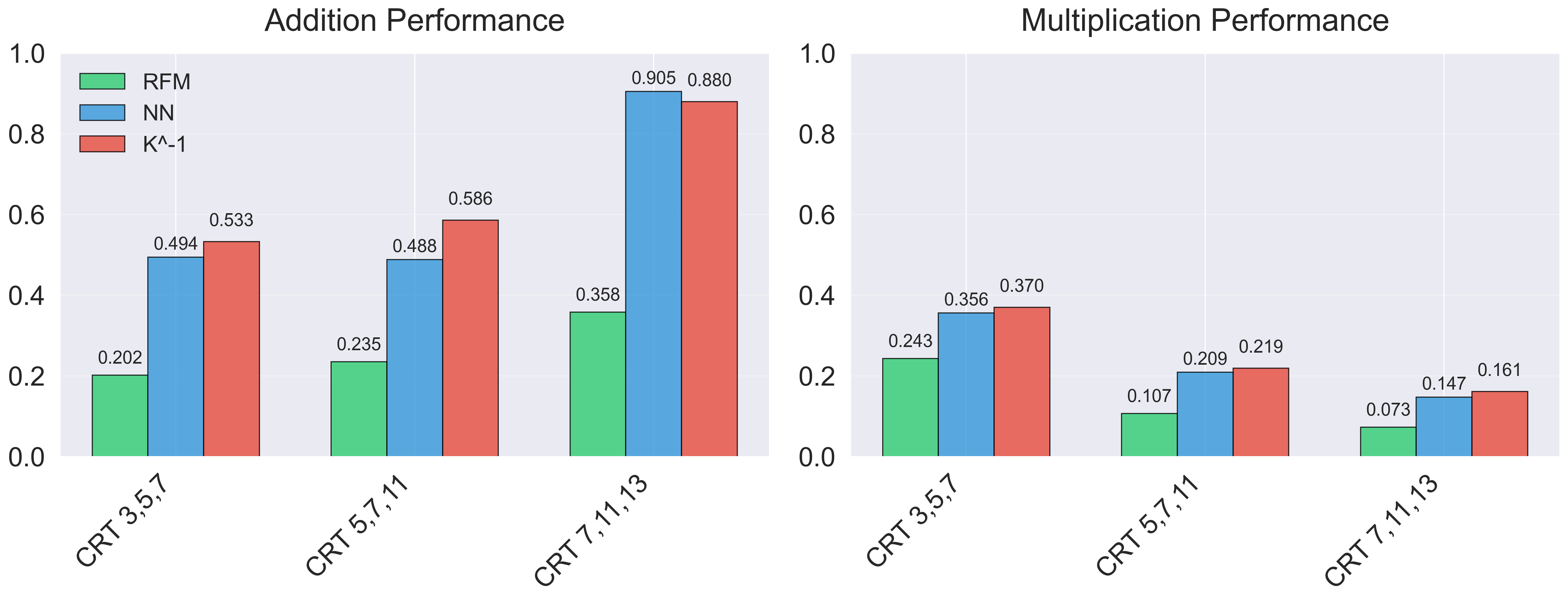}
    \end{subfigure}
    \caption[A comparison of RFMs, K-Inverse-RFMs, and neural networks across the different CRT encoding sizes given in \ref{crt-results}...]{A comparison of RFMs, K-Inverse-RFMs, and neural networks across the different CRT encoding sizes given in \ref{crt-results}. In 5/6 encoding-task pairs, the K-Inverse-RFM outperforms both the RFM and neural network.}
    \label{k_inverse_mr}
\end{figure}

We conclude our result set with a comparison between RFMs, K-Inverse-RFMs, and neural networks on the various CRT encodings introduced in Section 3.3. This scenario marks the most definitive success story of RFMs, as they surpass neural networks in five of six comparisons. We report these results in figure \ref{k_inverse_mr}.

While no feature learning occurs in either the neural network nor the RFM, the K-Inverse-RFM is able to attain superior performance by accomplishing a limited amount of feature learning. Unlike the neural network or the RFM, whose performances decrease/stagnate after the first epoch, K-Inverse-RFMs are able to improve performance through the first 5-20 epochs (depending on CRT coprimes). A power of $d=0.6$ is used for the actual K-Inverse-RFM formulation, as it performs a stabilizing reduction upon the extremely large features magnitudes given by a kernel on top of CRT inputs.

\section{Synthesis}
Holistically, our results show that K-Inverse-RFMs are more performant, sample efficient, and capable of learning generalizable features than RFMs. While in many scenarios a gap between K-Inverse-RFMs and neural networks still exists, it is now a far smaller gap, in some cases reduced by as much as 53\% in absolute test accuracy. 

Further, we discover that performance of both RFMs and K-Inverse-RFMs scales almost linearly with the number of unique points present in the data. The primary difference betwen the two is just that K-Inverse-RFMs have a superior scaling rate, likely as a result of the multiclass feature learning capabilities encouraged by their feature-space-to-feature-space mapping.

In some scenarios, particularly when decoding CRT data representations, K-Inverse-RFMs actually outperform neural networks. The reason could lie in the fact that - as we've seen from Laplace kernels trained on neural network features -  a kernel could be a more powerful predictor than several layers of a neural network. This suggests that an RFM that is able to learn quality features could actually outperform neural networks entirely. We leave elucidation and exploration of this phenomena to future work.

%% file: 5_conclusions_and_future_avenues.tex
\chapter{Conclusions, Limitations, and Future Avenues}
In this work, we explore the performance of RFMs across modular arithmetic and GCD tasks, discovering a gap between the normally performant RFMs and neural networks on noisy, unevenly distributed, and complexly represented data. We conduct a thorough analysis of these gaps and discover that they emerge due to the weaker feature learning capabilities of the RFM. To address these gaps, we construct the K-Inverse RFM. Our proposed model, as we showed in Chapter 4, is more performant, sample efficient, and generalizable. Further, it is very close in efficiency to the RFM itself and can also be easily trained on a CPU.

In our exploration, we find several interesting observations that could be of practical value and fuel future work. For one, we found that a Laplace kernel trained on post-nonlinearity first layer features often outperformed the neural network itself and seemed immune to overfitting. We also found that RFMs do not benefit from several instances of the same training point, and are far less impacted by imbalance in the dataset than they are by simple lack of unique training data. This provides a simple recipe for major performance improvement.

Future work could focus on extending these results to non-mathematical tasks. Showing generalization to important real world tasks in NLP, Computer Vision, and Tabular Benchmarks was beyond the scope of our work, and the generalization of these results to those domains remains unproven. Also, our work deals only with Feedforward neural network for simplicity, and the behavior of modern day modern day Transformers could be quite different. Proving both of these generalizations would address a major limitation of this work and allow practical use.

Finally, the K-Inverse RFM introduced in this work still lags behind neural networks in several scenarios, and a portion of the gap remains to be closed. Since the gap has been observed to be one of feature learning, exploring modification of the AGOP so as to enhance the K-Inverse RFMs feature learning capabilities could be a fruitful area of research.

%% file: A1_other_interesting_avenues_observed.tex
\chapter{Interesting Observations on the Nature of Neural Networks}
The phenomena observed in this work elucidate much regarding the inherent robustness of RFMs and neural networks to different data-level modifications. A particularly interesting phenomena, found outside the context of K-Inverse, was the surprising superb performance of kernels trained on top of neural net features. We perform a deeper exploration in the following sections, aiming to better understand the origins of this phenomena.

\section{Neural Networks Learn Robust Features Early}
Throughout our label noise experiments, we noticed that much of the RFM-NN gap was one of feature learning. We also find that the post-nonlinearity features of the first layer were often sufficient for a kernel machine to \textbf{improve} upon the results of the entire neural metwork, yielding some natural questions: How long does it take to really train these features? Could we quickly train neural network first layer features and then simply fit a kernel regressor on top of it? Are these features, like the neural network in its entirety, prone to overfitting? We answer all these questions, and more, with the experiments in this section.

To test the speed and robustness of feature learning, we first conduct the following experiment. We evaluate two tasks - modular addition and modular multiplication - at 32$\%$ label noise. For each of these tasks, we train three models:

\begin{enumerate}
    \item A base neural network with two hidden layers, a quadratic activation function, and hidden with of 512.
    \item A Laplace kernel using the post-nonlinearity first-layer features of the base neural network, evaluated at the relevant training step.
    \item A Laplace kernel using the post-nonlinearity second-layer features of the base neural network, evaluated at the relevant training step.
\end{enumerate}

We run training for 850 epochs and provide our results in the plot below:

\begin{figure}[h]
    \centering
    \begin{subfigure}{0.46\textwidth}
        \centering
        \includegraphics[width=\linewidth]{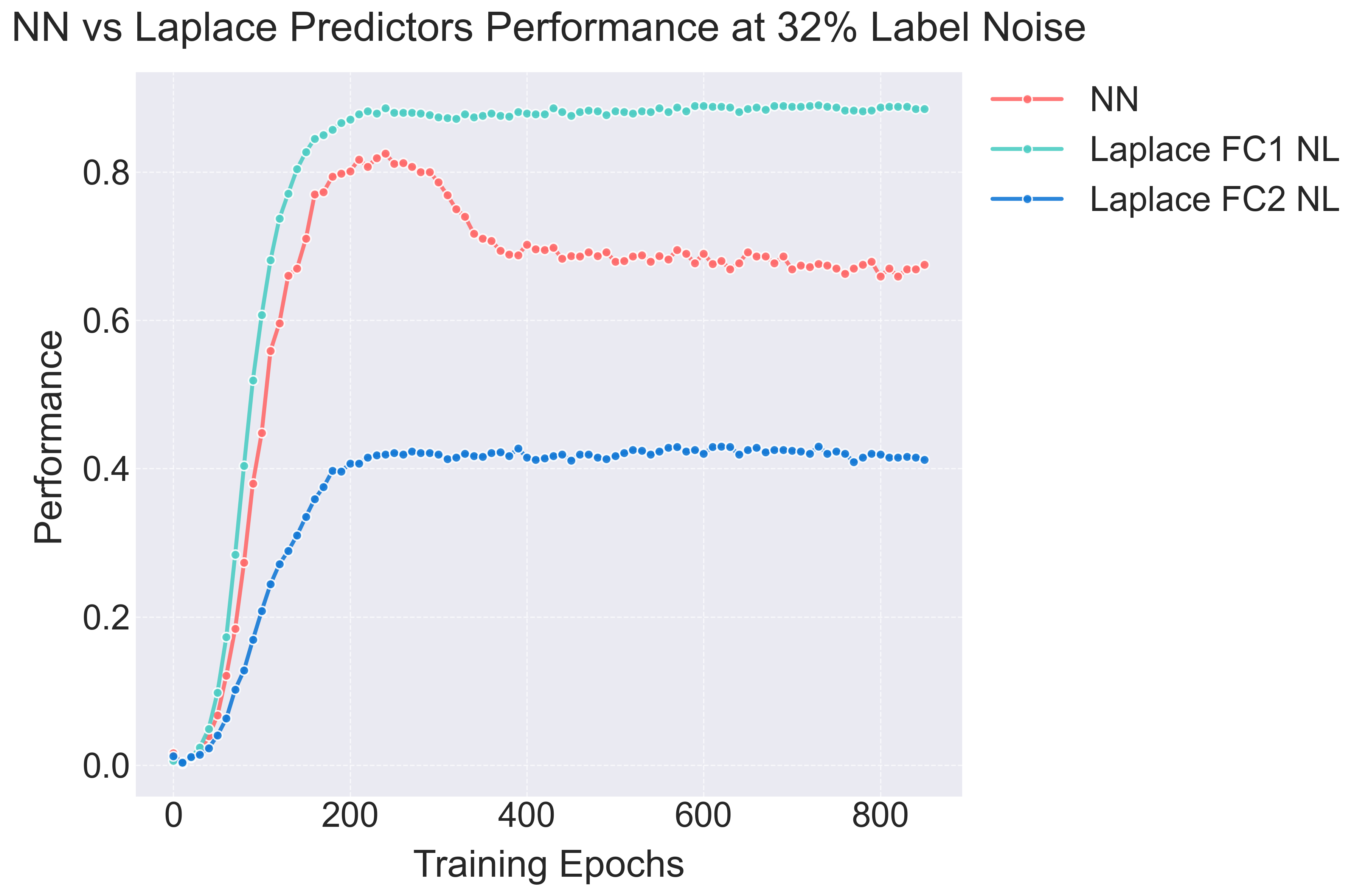}
    \end{subfigure}
    \begin{subfigure}{0.46\textwidth}
        \centering
        \includegraphics[width=\linewidth]{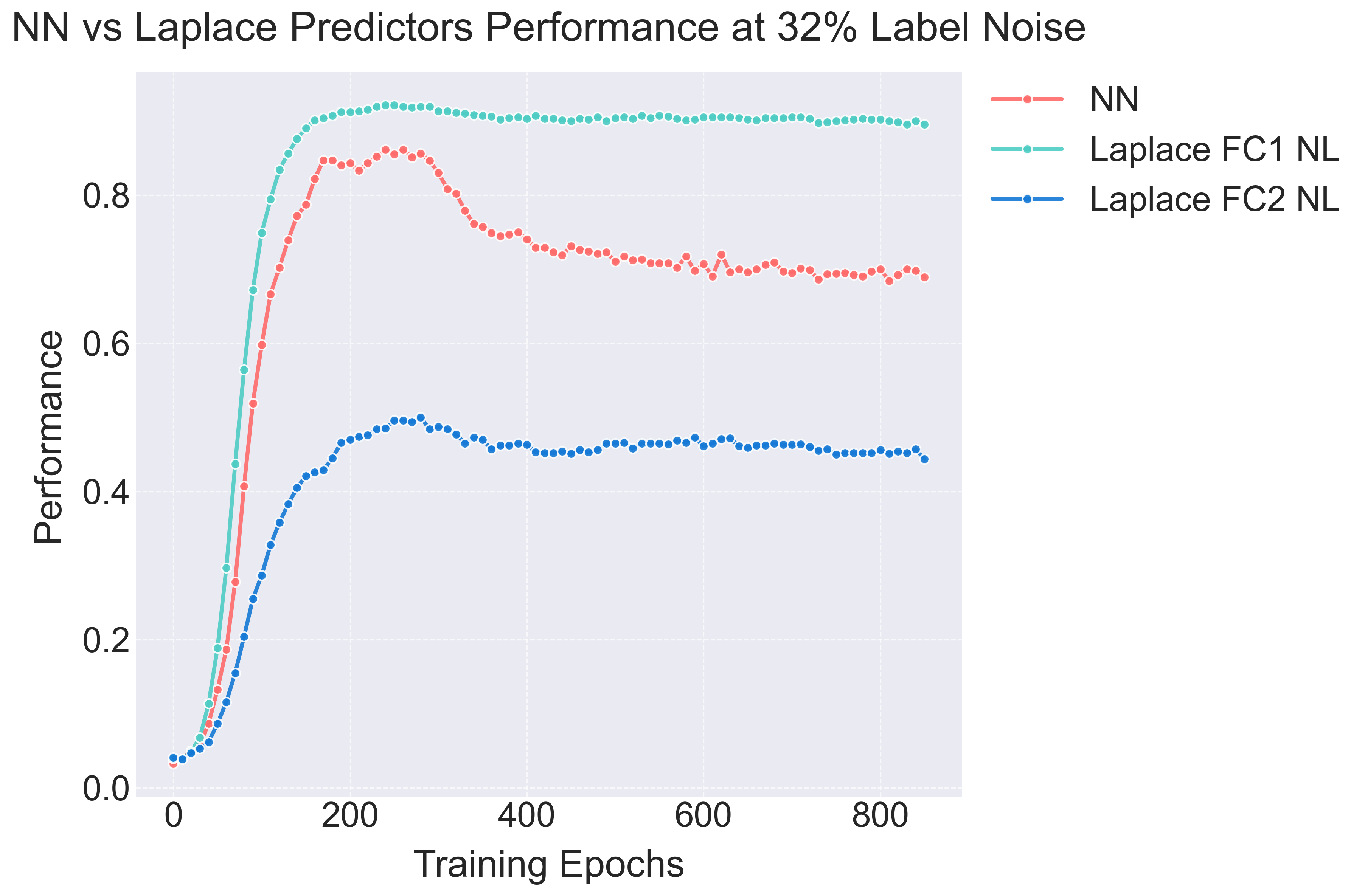}
    \end{subfigure}
    \begin{subfigure}{0.46\textwidth}
        \centering
        \includegraphics[width=\linewidth]{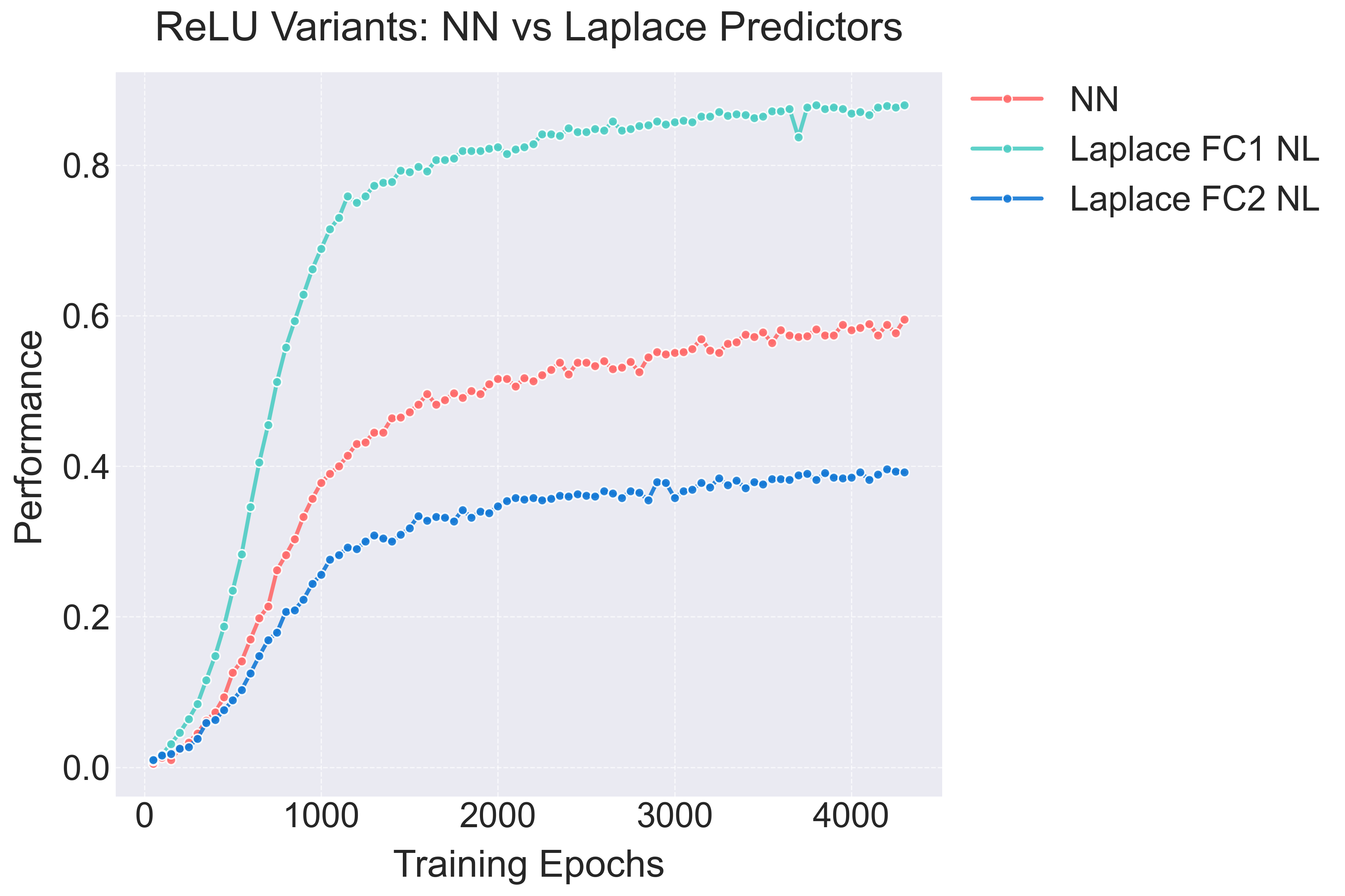}
    \end{subfigure}  
    \begin{subfigure}{0.46\textwidth}
        \centering
        \includegraphics[width=\linewidth]{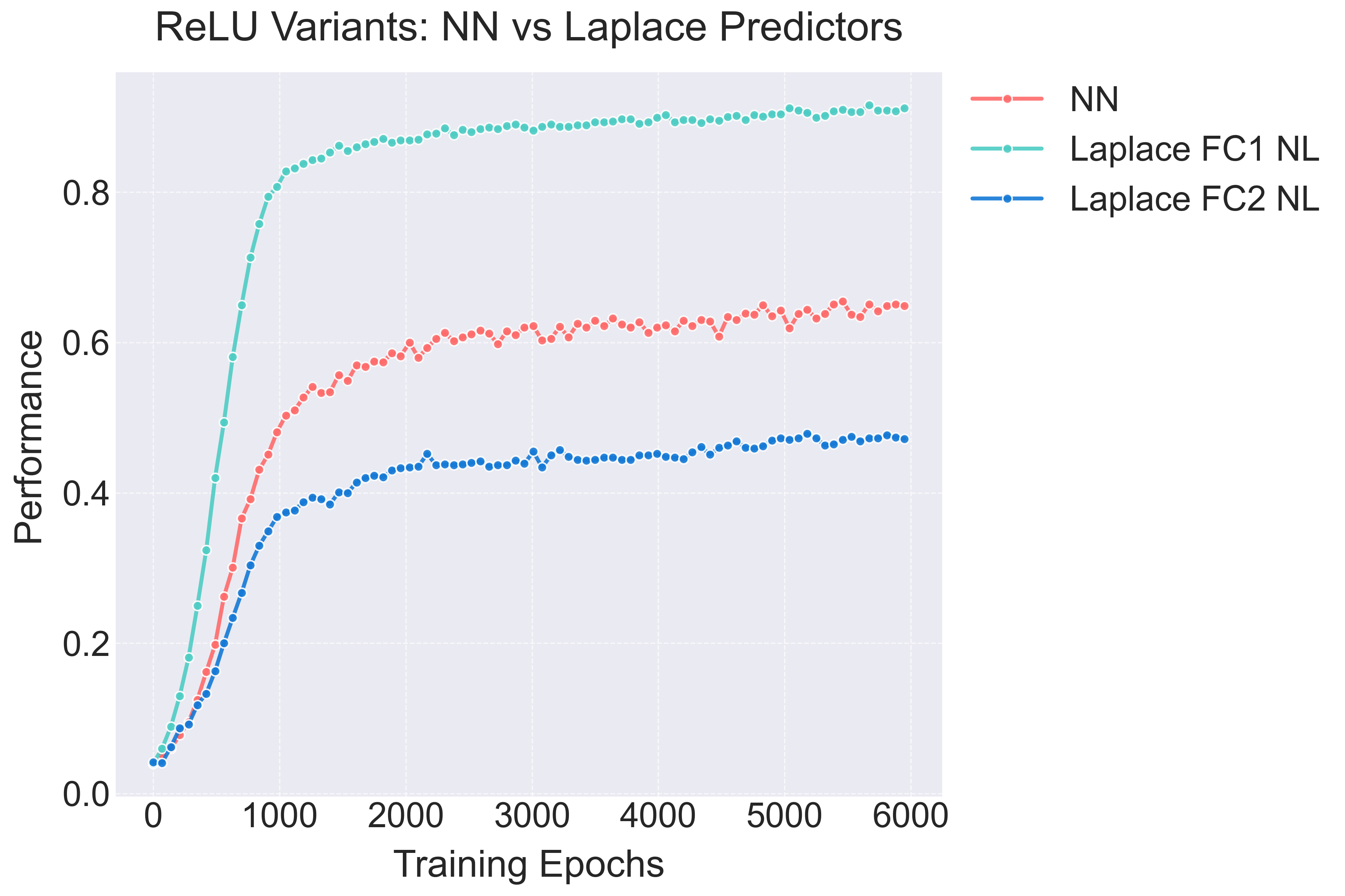}
    \end{subfigure}  
    \caption[textbf{Top row}: Neural networks trained at 32\% training set label noise with quadratic activation evaluated against Laplace kernel predictors trained on their first and second layer post-nonlinearity features...]{\textbf{Top row}: Neural networks trained at 32\% training set label noise with quadratic activation evaluated against Laplace kernel predictors trained on their first and second layer post-nonlinearity features. The left plot shows the task of modular addition, and the right plot that of modular multiplication. \textbf{Bottom row}: Neural networks trained at 32\% training set label noise with ReLU activation evaluated against Laplace kernel predictors trained on their first and second layer post-nonlinearity features. The left plot shows the task of modular addition, and the right plot that of modular multiplication.}
    \label{laplace_nn_ln_comp}
\end{figure}

Interestingly enough, our results indicate two items with relative clarity. Firstly, as seen from the FC1 post-nonlinearity features in light blue, using a Laplace kernel as a predictor is more powerful than both the second layer and the linear prediction head combined, as the Laplace kernel with FC1 post-nonlinearity features outperforms neural networks in their entirety. The difference is particularly striking with a ReLU activation function, where at certain points in training we can see test set accuracy differences measuring as much as 30\%. These features are also immune to overfitting, as while the neural network (denoted in red) overfits and its performance tends downward, the features of the first layer continue to improve when coupled with the Laplace predictor. They also significantly outperform second layer features and thus seem to be more general, corroborating existing literature \cite{deeplearningexampledepth}.

Further, while the post non-linearity second layer features don't perform as well, their performance \textbf{does not worsen} as training occurs, even as the neural network overfits. This could attest to the problem not being the overfitting of intermediate features, but rather the overfitting of the linear prediction head. Such findings could have important implications, as they suggest that it may be possible to train a neural network without fear of overfitting, extract the relevant features, and use a carefully chosen predictor to yield strong results.

\section{Second Layer is for Specificity}
In the previous section, we attained a stark improvement when using a Laplace kernel to replace the entire second layer + linear predictor of a neural network. Such results may naturally cause one to question the importance of the second layer altogether for feature learning on discrete mathematical tasks. After all, if a kernel mechanism that is largely incapable of learning multiclass features is able to use first layer representations to outperform a neural network in its entirety, perhaps it may be the case that the first layer learns all relevant features for this task. In this instance, the second layer may simply be memorizing specific examples and hurting the generalization of the neural network altogether.

To test whether this might be the case in the context of modular arithmetic, we experiment with increasing the width of the second layer of the network. One justification for this approach is that if the second layer truly isn't useful in feature learning, we would expect to see no significant improvement from increasing its richness. Perhaps, we might even see a decay as a result of increased overfitting.

\begin{figure}[h]
    \centering
    \begin{subfigure}{0.46\textwidth}
        \centering
        \includegraphics[width=\linewidth]{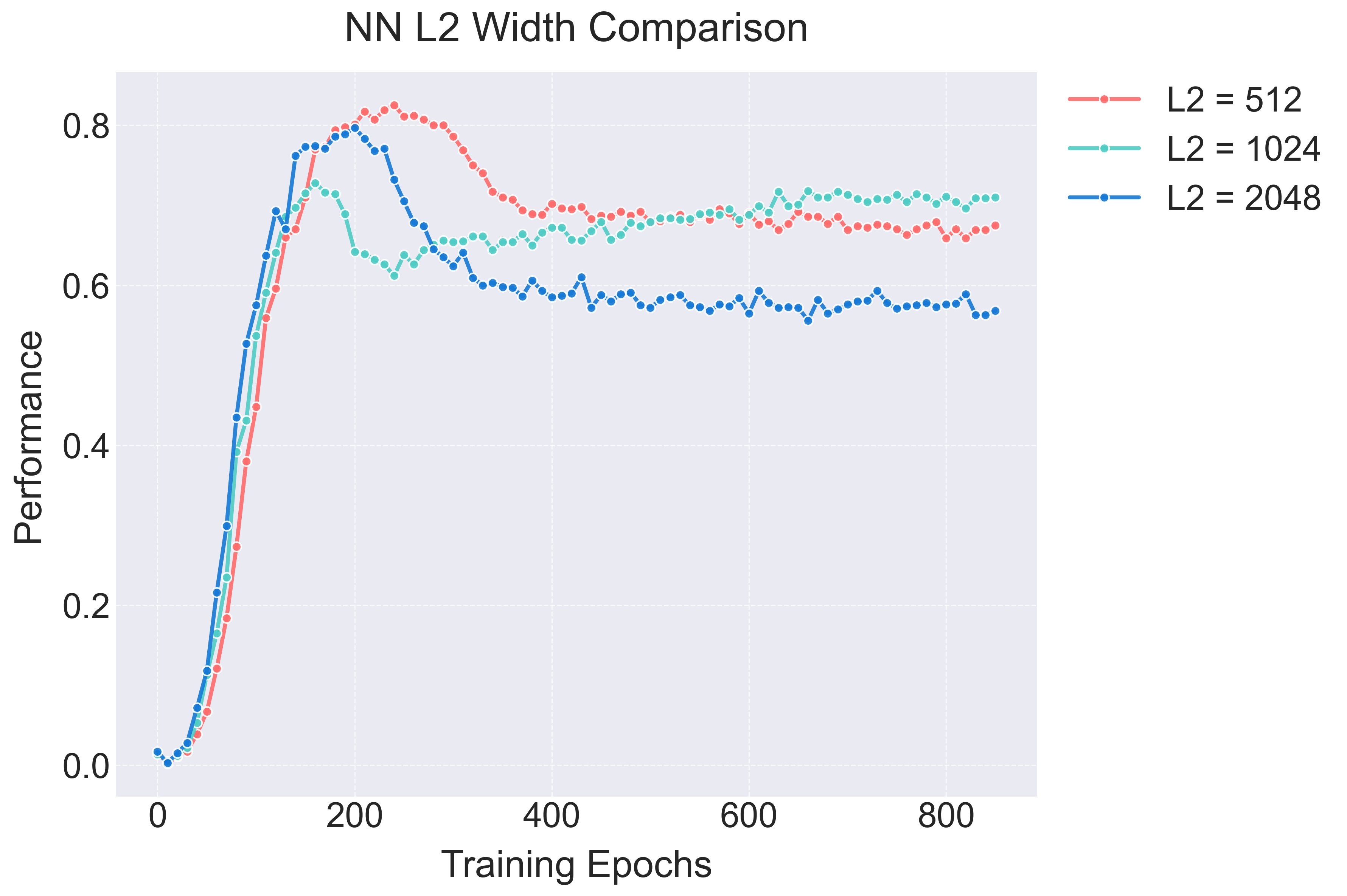}
    \end{subfigure}
    \begin{subfigure}{0.46\textwidth}
        \centering
        \includegraphics[width=\linewidth]{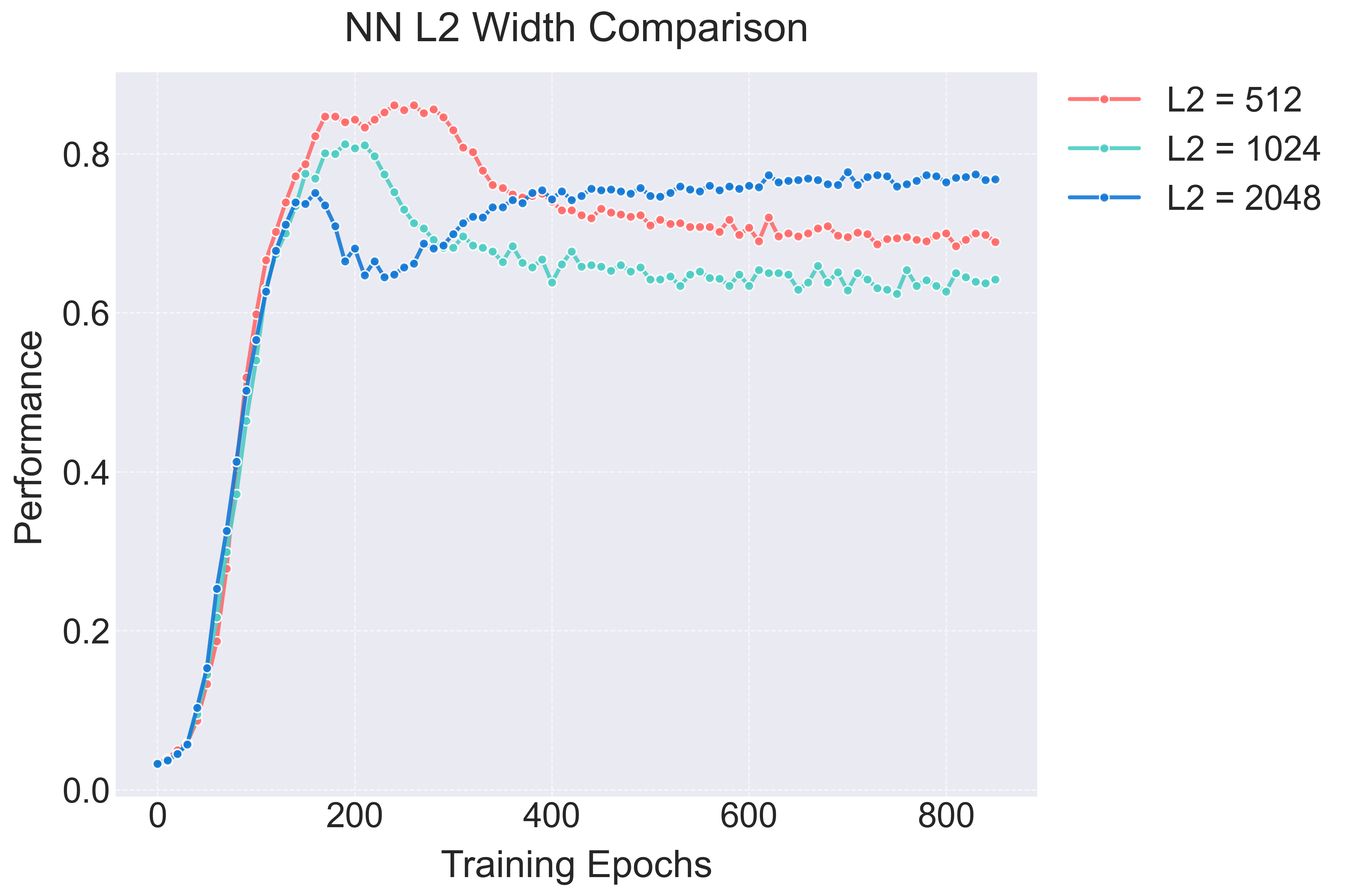}
    \end{subfigure}  
    \caption[Performance of a 2-layer neural network with a quadratic activation function at different second layer widths...]{Performance of a 2-layer neural network with a quadratic activation function at different second layer widths. The tasks are 32\% label noise modular addition (left) and multiplication (right) }
    \label{l2_width_comp}
\end{figure}

Indeed, this is precisely what we see in our experiment in \ref{l2_width_comp}. Our results show that there isn't an inherent advantage to increasing the second layer width - increasing its feature learning capacity is not helpful for the task. In fact, our results suggest that increasing the width may yield greater overfitting, perhaps to the 32\% of label noise present in our modular arithmetic data.

\section{The Relationship of Width and Role}
\begin{figure}[h]
    \centering
    \begin{subfigure}{0.46\textwidth}
        \centering
        \includegraphics[width=\linewidth]{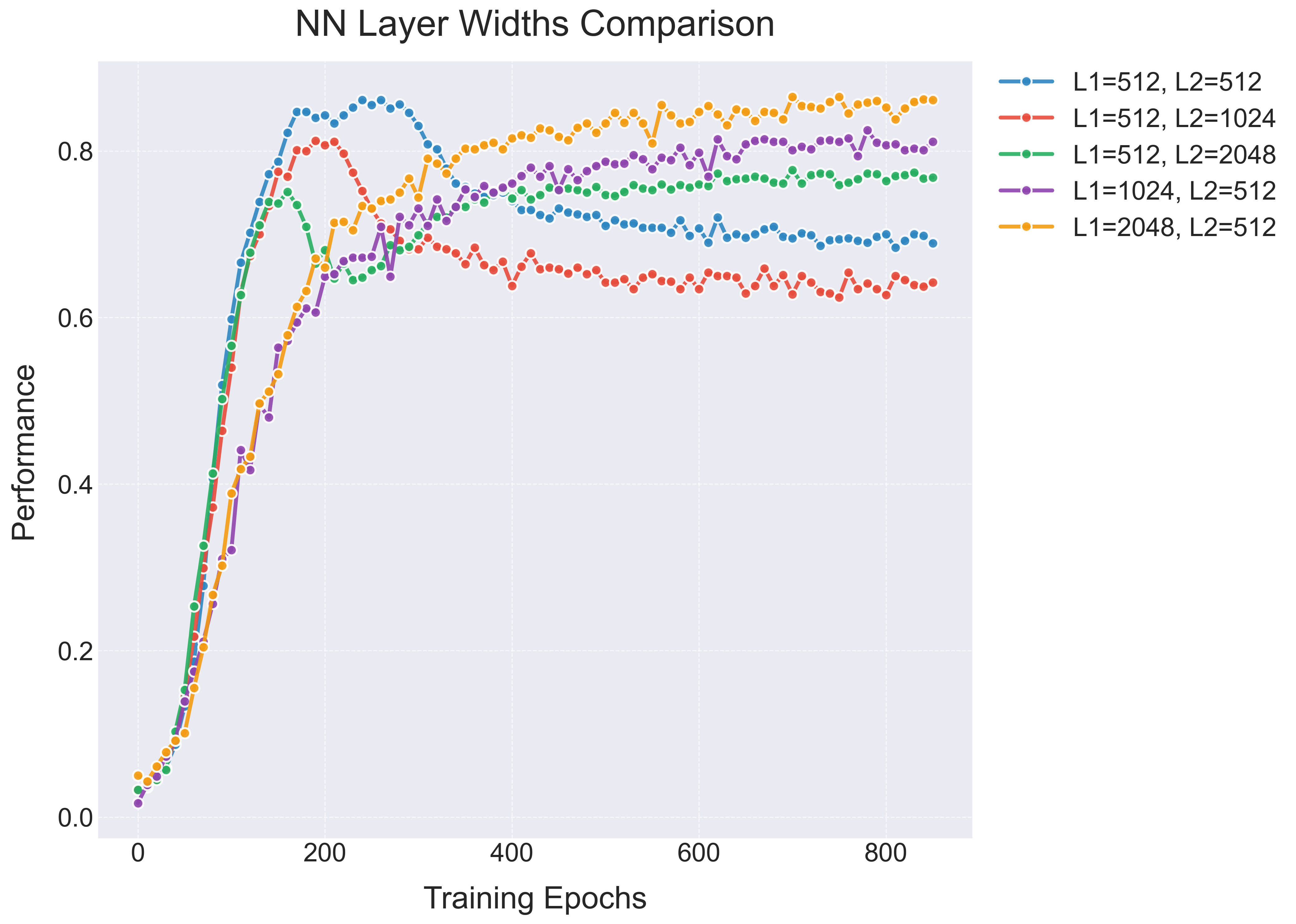}
    \end{subfigure}
       \begin{subfigure}{0.46\textwidth}
        \centering
        \includegraphics[width=\linewidth]{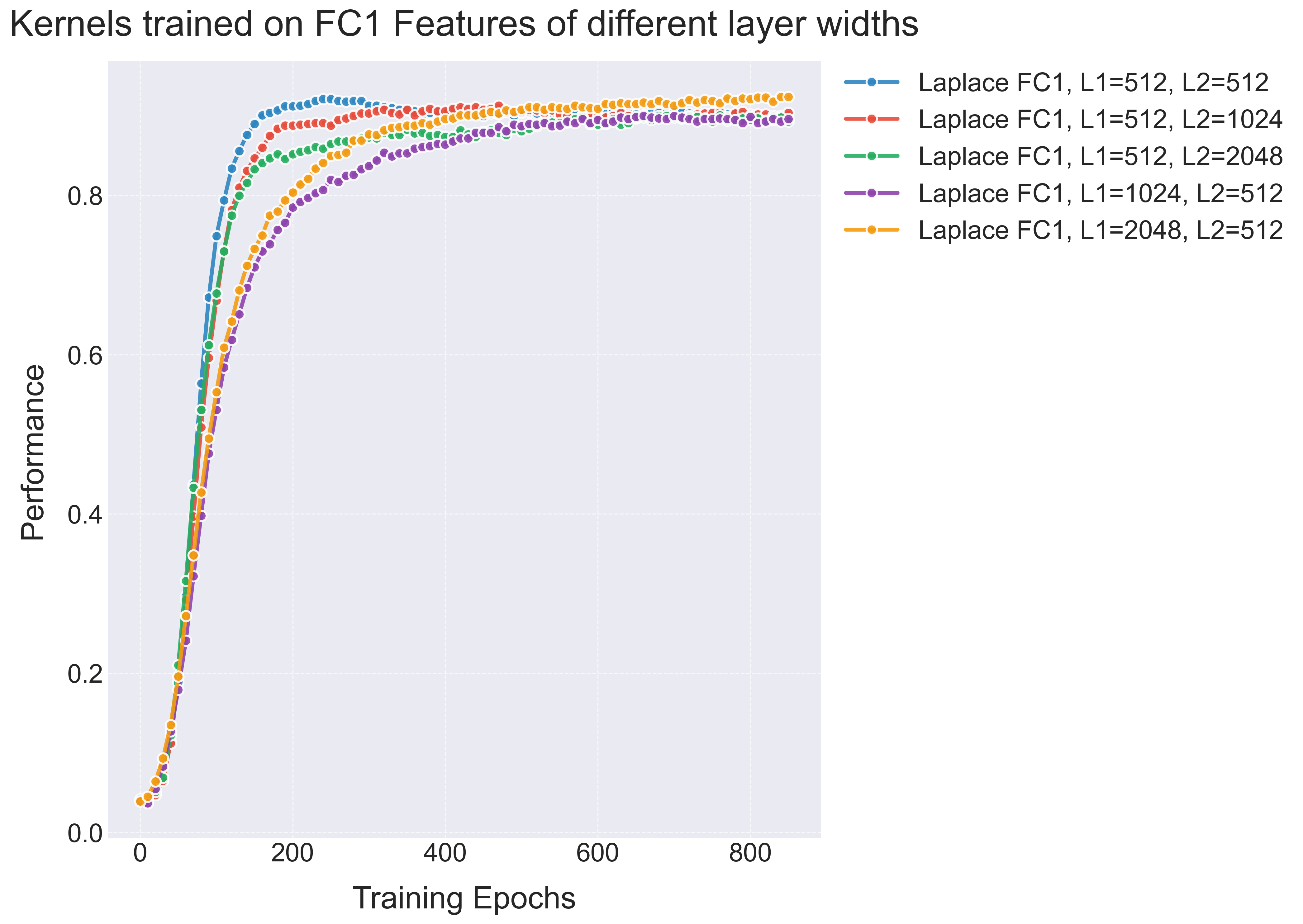}
    \end{subfigure}
    \begin{subfigure}{0.46\textwidth}
        \centering
        \includegraphics[width=\linewidth]{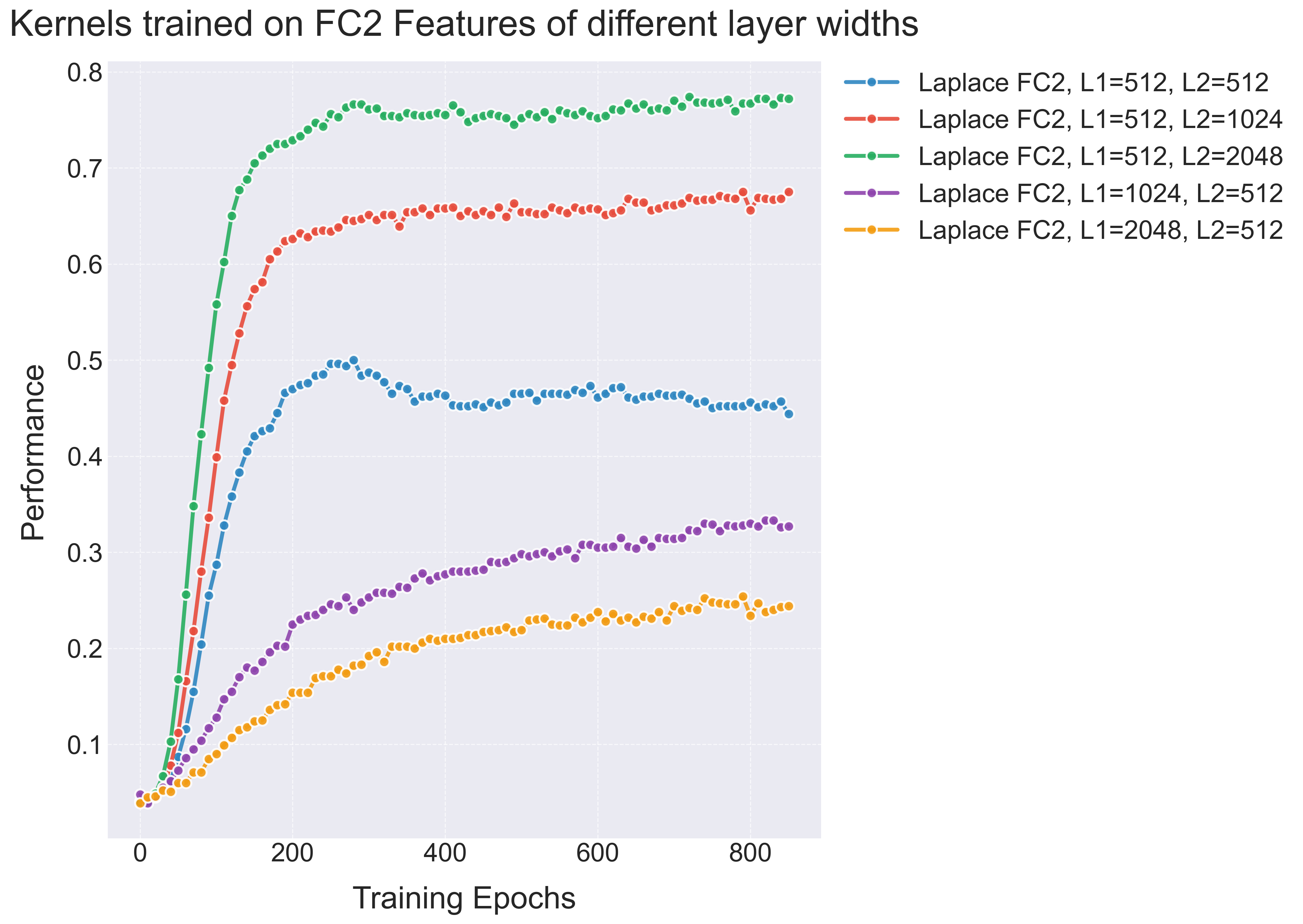}
    \end{subfigure}  
    \caption[Performance of a 2-layer neural network with a quadratic activation function with varying first and second layer widths on modular multiplication...]{Performance of a 2-layer neural network with a quadratic activation function with varying first and second layer widths on modular multiplication. \textbf{Left}: variants with the largest first layer widths perform the best, showing the importance of the first layer in feature learning. \textbf{Right}: varying the neural network architecutre has little effect on the richness of features from the first layer, as shown by similar performance irrespective of architecture. \textbf{Bottom}: the effect of varying the first and second layer size on the features learned by the second layer. The second layer learns richer features when it is far larger than the first layer.}
    \label{width_variants}
\end{figure}

While our results show that there is no inherent advantage to increasing the size of the second layer, might the same be said for the first layer? Perhaps a larger first layer may also hinder learning. To test this hypothesis and complete the picture we've painted in this appendix, we vary the size of the first layer and report results. Notably, we observe that the best performing neural networks are precisely those with the largest first layers, bolstering the hypothesis that for modular arithmetic the first layer is the one crucial for feature learning. We present our results in \ref{width_variants}.

We conclude our experiments for this section with a final check as to the effect of varying first and second layer widths on the richness of features, as measured by the performance of a Laplace kernel that takes these features as inputs. Interestingly, the ratio of neurons in the first to the second layer significantly influences the quality of features learned in the second layer, as shown in the bottom plot of Figure \ref{width_variants}. In contrast, the feature performance in the first layer remains very stable across different neuron ratios. This may indicate that deeper layers must be disproportionately large computational workhorses to capture useful representations for modular arithmetic. However, as demonstrated earlier, this additional capacity might not yield performance gains, since first layer alone learns the necessary features. Ultimately, these experiments may suggest that intelligently distributing the compute ratio in favor of earlier layers to develop a reliance on generalizable features may be a useful method for model improvement.